\documentclass[lettersize,journal]{IEEEtran}

\usepackage{amsmath,amsfonts,amssymb}
\usepackage{algorithmic}
\usepackage{algorithm}
\usepackage{array}
\usepackage[caption=false,font=normalsize,labelfont=sf,textfont=sf]{subfig}
\usepackage{textcomp}
\usepackage{stfloats}
\usepackage{graphicx}
\usepackage{cite}
\usepackage{tabularx}
\usepackage{booktabs}
\usepackage{makecell}
\usepackage{multirow}
\usepackage[table]{xcolor}
\usepackage{colortbl}
\usepackage{url}
\usepackage[hidelinks]{hyperref}
\hypersetup{
  pdftitle={Variance-Guided Spatial Attention Fusion for Robust End-to-End Driving under Asymmetric Sensor Degradation},
  pdfauthor={Weizhi Tao, Zengwang Jin, Xiao Wang, Hailong Huang},
  pdfsubject={Preprint submitted to IEEE for possible publication},
  pdfkeywords={Autonomous vehicles, vehicular automation, multimodal sensor fusion, uncertainty estimation, sensor degradation, robust end-to-end driving}
}

\begin{document}

\title{Variance-Guided Spatial Attention Fusion for Robust End-to-End Driving under Asymmetric Sensor Degradation}

\author{Weizhi~Tao, Zengwang~Jin, Xiao~Wang,~\IEEEmembership{Senior Member,~IEEE}, and Hailong~Huang,~\IEEEmembership{Senior Member,~IEEE}%
\thanks{Weizhi Tao and Hailong Huang are with the Department of
Aeronautical and Aviation Engineering, The Hong Kong Polytechnic
University, 999077, Hong Kong, China (e-mail:
weizhi.tao@connect.polyu.hk; hailong.huang@polyu.edu.hk).}%
\thanks{Zengwang Jin is with the School of Robotics, Anhui University,
Hefei 230039, China; with the Anhui Provincial Engineering Research
Center for Unmanned System and Intelligent Technology, Anhui University,
Hefei 230039, China; and also with the Shenzhen Research Institute of
Northwestern Polytechnical University, Shenzhen 518057, China (e-mail:
zwjin@ahu.edu.cn).}%
\thanks{Xiao Wang is with the School of Robotics and Institute of
Embodied Intelligence, Anhui University, Hefei 230039, China (e-mail:
xiao.wang@ahu.edu.cn).}%
\thanks{Corresponding author: Hailong Huang.}%
\thanks{This work has been submitted to the IEEE for possible publication.
Copyright may be transferred without notice, after which this version may
no longer be accessible.}%
}

\markboth{Preprint}{Tao
\MakeLowercase{\textit{et al.}}: Variance-Guided Spatial Attention Fusion
under Asymmetric Sensor Degradation}

\maketitle

\begin{abstract}
End-to-end multimodal driving has progressed rapidly by fusing camera
and LiDAR streams. Existing pipelines remain fragile under
\emph{asymmetric sensor degradation}, where either an entire modality or
only a localized region is corrupted while other regions remain useful.
The key difficulty is not simply to add an uncertainty head, but to
obtain dense reliability supervision, calibrate this reliability against
physical fault severity, and use it before unreliable features bias the
planner. We propose Variance-Guided Spatial Attention Fusion (VG-SAF),
in which dense heteroscedastic reliability estimates act as interpretable
spatial gates. The framework couples three components. First, a physically
grounded augmentor simulates representative camera and LiDAR failures and
emits a continuous spatial mask, providing dense supervision without
additional annotation. Second, modality-specific experts predict per-pixel
reliability scales through cross-branch dense distillation in log space,
enforcing a monotone severity-to-scale response. Third, calibrated
reliability maps drive a hybrid attention mechanism that suppresses
unreliable cells with a local spatial gate and arbitrates between modalities
through a cross-modal trust softmax. A Laplace uncertainty head emits a
systemic waypoint uncertainty scale that signals severe or combined sensor
degradation, including severities outside the training ranges. On the CARLA
Longest6 benchmark, VG-SAF consistently improves closed-loop robustness
over the baselines across camera-only, LiDAR-only, and joint degradation
regimes, as measured by driving score, route completion, and infraction
score.
\end{abstract}

\begin{IEEEkeywords}
Autonomous vehicles, vehicular automation, multimodal sensor fusion,
uncertainty estimation, sensor degradation, robust end-to-end driving.
\end{IEEEkeywords}

\section{Introduction}\label{sec:intro}

\IEEEPARstart{F}{ully} autonomous driving is one of the defining
pursuits of contemporary artificial intelligence, with broad
implications for urban mobility and traffic safety
\cite{cite:chen_survey}. Sustained progress on
benchmarks such as nuScenes \cite{cite:nuscenes}, the CARLA
Leaderboard \cite{cite:carla, cite:leaderboard} and Longest6
\cite{cite:fusion_transformer} has produced end-to-end policies that
map raw sensor input directly to control. Recent studies on intelligent
vehicular systems approach this goal from complementary angles, including
hierarchical reinforcement learning, general/specialized driving-model
interfaces and resource modeling for embedded driving intelligence
\cite{cite:tvt_vision_hrl, cite:tvt_netroller, cite:tvt_multiruler}.
The common requirement behind these advances is reliable information
fusion: the planner must receive perception cues that remain useful when
the operating condition changes. Camera, LiDAR and radar provide
complementary appearance, geometry and ranging cues
\cite{cite:chen_survey, cite:tvt_openmpd, cite:tvt_radar_lidar}, so
multimodal sensing is widely adopted as the foundation of autonomous
vehicles. Therefore, robust fusion is directly tied to route completion,
infraction reduction and operational continuity in intelligent vehicular
systems.

\begin{figure}[!t]
\centering
\includegraphics[width=\columnwidth]{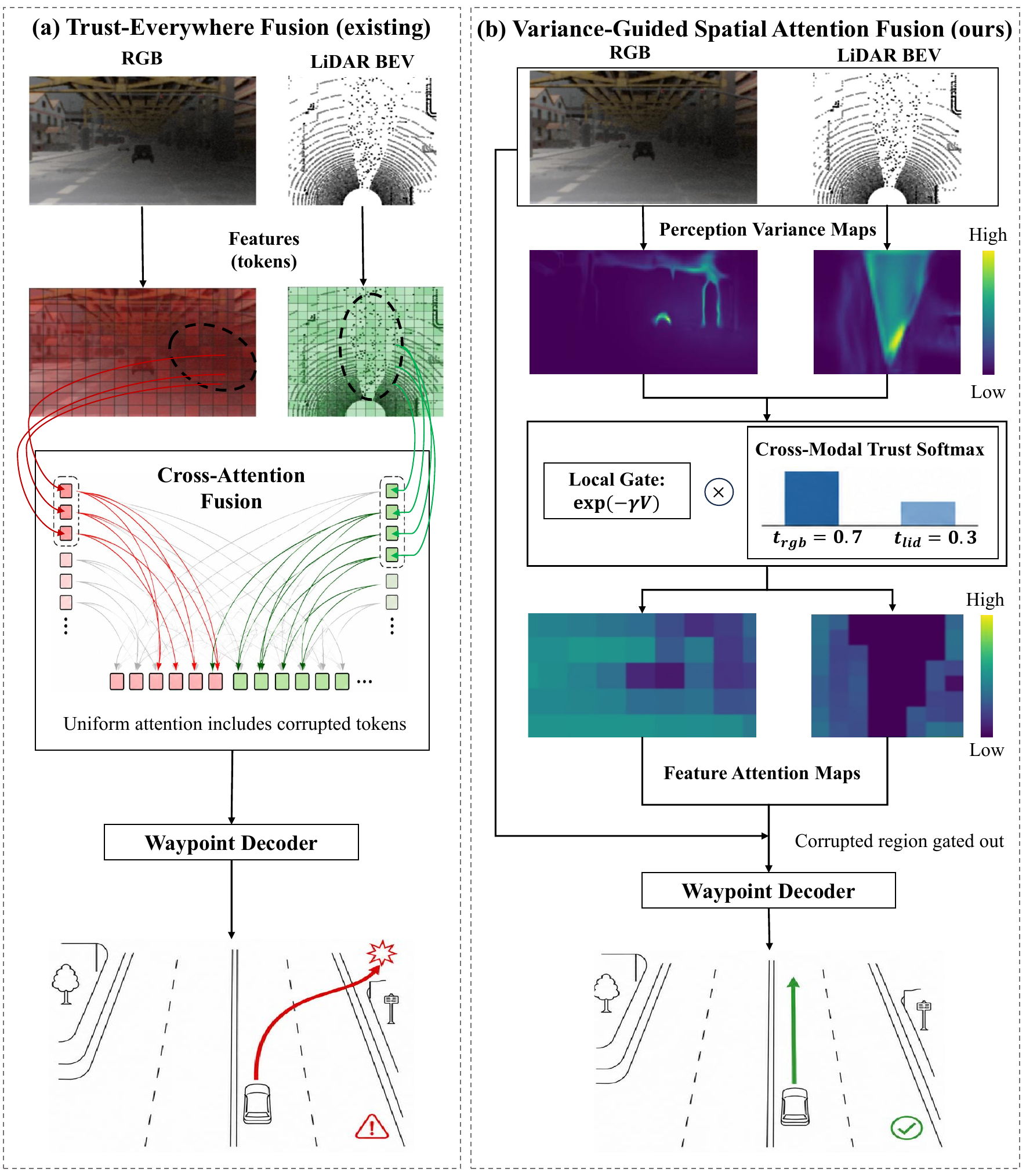}
\caption{Trust-everywhere fusion versus the proposed variance-guided
spatial attention fusion (VG-SAF) under sensor corruption. (a) A
cross-attention stage admits every token regardless of its sensor-side
quality. The damaged region (dashed ellipse) propagates uniformly into
the fused representation and biases the planner. (b) VG-SAF estimates
a dense variance map for each modality. A local spatial gate
$\exp(-\eta^{\mathrm{loc}}_m\bar V_m)$ and a cross-modal trust softmax suppress the
corrupted region while preserving the surviving modality.}
\label{fig:intro_concept}
\end{figure}

However, the complementarity holds only when both sensor streams are
nominal, i.e., when the camera and LiDAR operate under normal sensor
health. Robustness benchmarks document that real-world failures are not
symmetric \cite{cite:imagenetc, cite:robo3d, cite:robobev}. They may be
spatially localized and asymmetric across modalities. A water droplet
corrupts a small patch of camera pixels, whereas a LiDAR blockage may
empty one angular frustum of the BEV scan. The safe response is therefore
not to discard an entire modality, but to discount the corrupted
\emph{cells} of a stream while continuing to use the \emph{remaining
cells} of the same stream. Existing intermediate-fusion architectures
\cite{cite:transfuser, cite:fusion_transformer, cite:interfuser,
cite:tcp, cite:thinktwice} and BEV-fusion architectures
\cite{cite:bevfusion_mit, cite:uniad, cite:vad} do not provide this
property. Their fusion weights are conditioned on the \emph{content} of
the features, not on the physical reliability of the cells that produced
them. Local corruption is therefore smeared through the attention map
rather than isolated to the cells that carry the corrupted physical
signal, which is precisely the resolution needed for safe local
maneuvers.

Robustness research in the domain has predominantly addressed a
modality-level threat model \cite{cite:unibev, cite:metabev,
cite:cafuser, cite:crossfuser, cite:maskfuser, cite:policyfuser}.
Such methods are effective when a complete stream is missing or when a
scene-level condition changes globally, but they are less suited to
localized asymmetric faults. A modality-level dropout decision either
keeps the damaged stream, allowing corrupted cells to enter the fused
representation, or suppresses the whole stream, discarding unaffected
cells that may still contain lane, obstacle or traffic-light evidence.
A global trust scalar or a scene-level condition token faces the same
resolution mismatch: it can summarize the environment but cannot select
which spatial cells of the modality are reliable. The machinery that
could in principle supply a per-cell trust signal, namely heteroscedastic
uncertainty estimation \cite{cite:kendall_gal, cite:gawlikowski_survey},
has been used mainly as a downstream diagnostic
\cite{cite:michelmore, cite:loquercio, cite:kendall_multitask}.
Existing methods have not explicitly used a mask-calibrated dense
variance as a \emph{structural} input for cross-modal fusion, leaving the
planner without a calibrated reliability field.

These observations expose three coupled challenges. First, dense
reliability supervision is needed at the cell level, but manual
uncertainty labels are unavailable in ordinary driving datasets. Second,
heteroscedastic loss attenuation alone relates the predicted scale to
supervised loss magnitude, which can be unstable under corrupted inputs
because severely damaged observations may no longer contain sufficient
evidence for reliable direct supervision. Third, the resulting reliability field must serve two
roles simultaneously: local spatial suppression inside each modality and
cross-modal arbitration between modalities. Fig.~\ref{fig:intro_concept}
contrasts the proposed mechanism with the cross-attention paradigm. We
promote heteroscedastic per-pixel predictive variance from a downstream
diagnostic to an architectural input of the fusion stage. A physically
grounded augmentor injects representative camera and LiDAR faults and,
alongside every augmented frame, emits a continuous corruption mask
$\mathbf{M}$ that supplies dense reliability supervision without
additional human annotation. The mask supervises cross-branch dense
distillation, which calibrates predicted log-variance against physical
fault severity rather than using only the residual proxy. The calibrated
variance then drives a hybrid attention that decouples within-modality
spatial gating from sample-level cross-modal trust. A Laplace uncertainty head emits a systemic waypoint uncertainty scale
that increases under severe or combined degradation and can be exposed to
a downstream safety controller.

The principal contributions of this paper are as follows.
\begin{enumerate}
\item We promote heteroscedastic per-pixel predictive variance from a
downstream diagnostic to a \emph{structural} input of multimodal fusion
attention for reliability-aware planning.
\item We introduce a physically grounded fault-injection augmentor
for the camera and the LiDAR streams. Every augmentation is paired
with a continuous, per-cell corruption mask. A four-phase
progressive curriculum uses the mask to calibrate the variance
heads against physical severity while preserving supervision on reliable
regions.
\item We design a cross-branch dense distillation loss that enforces
a monotone, log-space alignment between predicted log-variance and
corruption severity, without auxiliary manual labels or separate
uncertainty annotations.
\item We design a hybrid variance-guided attention that decouples
per-cell spatial gating from sample-level cross-modal arbitration.
A Laplace uncertainty head exposes a systemic waypoint uncertainty
scale that responds to severe degradation and to severity extrapolation
beyond the training ranges, providing an interpretable safety signal.
\end{enumerate}
We evaluate VG-SAF on the CARLA Longest6 benchmark under camera-only,
LiDAR-only and joint corruption regimes. The evaluation reports the
standard closed-loop driving score, route completion and infraction
score, and the results show that variance-guided fusion consistently
improves robustness over feature-only cross-attention baselines while
maintaining interpretable modality trust and per-cell attention maps.

The remainder of the paper is organized as follows.
Section~\ref{sec:related} surveys the three research threads on which
this work draws. Section~\ref{sec:method} presents the full framework,
including the mask-weighted supervision, confidence-aware consistency,
cross-branch dense distillation and hybrid variance-guided fusion.
Section~\ref{sec:experiments} reports the closed-loop, ablation and
qualitative evaluations. Section~\ref{sec:limitations} discusses three
limitations and future directions, and Section~\ref{sec:conclusion}
concludes the paper from the perspective of robust vehicular autonomy.

\section{Related Work}\label{sec:related}

\subsection{Multimodal End-to-End Driving and Cross-Modal Fusion}
End-to-end driving has progressed from conditional imitation
learning \cite{cite:cil, cite:cilrs, cite:lbc} to multimodal
architectures that align RGB and LiDAR streams through shared latent
representations and cross-attention. Recent work also studies
end-to-end vehicular autonomy from complementary perspectives:
hierarchical reinforcement learning for vision-based driving
\cite{cite:tvt_vision_hrl}, general/specialized model interfacing for
end-to-end driving \cite{cite:tvt_netroller}, and resource modeling for
embedded autonomous-driving intelligence \cite{cite:tvt_multiruler}.
A representative early multimodal instance is TransFuser
\cite{cite:transfuser, cite:fusion_transformer}, which interconnects
two convolutional encoders through multi-resolution cross-attention.
Subsequent architectures explore variations within the same paradigm.
InterFuser \cite{cite:interfuser} adds dense object prediction together
with a learned safety filter. TCP \cite{cite:tcp} couples trajectory
and direct control prediction. ThinkTwice \cite{cite:thinktwice}
refines the BEV through a coarse-to-fine waypoint planner.
A second strand fuses modalities in a shared bird's-eye-view space.
BEVFusion~\cite{cite:bevfusion_mit, cite:bevfusion_pku} lifts camera
features into the BEV through depth-distribution estimation and
aligns them with LiDAR voxels under a unified representation.
Earlier cross-modal works fuse LiDAR and camera at the point or
voxel level, as in PointPainting \cite{cite:pointpainting} and the
continuous-fusion family \cite{cite:contfuse, cite:mmf}. OpenMPD provides
a multimodal autonomous-driving perception benchmark, and radar--LiDAR
fusion further illustrates the benefit of combining
sensors with complementary failure characteristics
\cite{cite:tvt_openmpd, cite:tvt_radar_lidar}.
UniAD~\cite{cite:uniad} and VAD~\cite{cite:vad} unify detection,
tracking, mapping and planning under a single transformer that
propagates features along the prediction path. A complementary line
scales the planner through unified transformers and latent
world-model representations \cite{cite:drivetransformer, cite:law}.
Across this body of work the fusion weights are conditioned on
\emph{features}, not on the physical reliability of the cells from
which those features were originally extracted, regardless of the
local sensor health at inference time.

\subsection{Uncertainty Quantification in Deep Networks}
Modern uncertainty quantification builds on the aleatoric/epistemic
taxonomy of \cite{cite:kendall_gal}. The aleatoric component is
heteroscedastic across the input and observable through a single
forward pass. The epistemic component can signal model uncertainty under distribution
shift. Stochastic forward passes \cite{cite:mcdropout,
cite:deepensembles} and the higher-order posteriors of evidential
deep learning \cite{cite:edr, cite:edl} approximate the epistemic
posterior. Their cost makes them unattractive for self-driving
latency budgets. Sampling-free heteroscedastic predictors
\cite{cite:kendall_gal, cite:gawlikowski_survey} predict per-pixel
negative log-likelihoods through a single forward pass at no
additional stochastic-sampling overhead, and deterministic variants
further compress the cost while preserving calibration \cite{cite:ddu}.
In autonomous driving, predicted uncertainty has been used primarily
as a downstream signal: as a crash predictor \cite{cite:michelmore}, a distribution-shift detector
or adapter \cite{cite:loquercio, cite:filos}, or a multi-task loss balancer
\cite{cite:kendall_multitask}. These works establish that uncertainty can
indicate reliability, but their uncertainty estimates are consumed after
the prediction has already been formed. By contrast, robust multimodal
fusion requires the reliability estimate to intervene \emph{before} the
corrupted features are fused. None of these methods treats per-cell
variance as a structural input to a multimodal fusion mechanism. The
role of the variance map as an upstream structural input to cross-modal
arbitration has therefore remained largely unexplored to date.

\subsection{Robust Multimodal Driving under Sensor Corruption}
Systematic robustness benchmarks include ImageNet-C
\cite{cite:imagenetc} on the image side and Robo3D
\cite{cite:robo3d} and RoboBEV \cite{cite:robobev} on the
3D-perception side. Physics-grounded simulators reproduce fog and
snowfall on real LiDAR scans
\cite{cite:lidar_fog, cite:lidar_snow}, and adversarial-robustness
benchmarks generate corruption-rich driving sequences in CARLA
\cite{cite:carla_gear}. Architectural robustness in fusion has been
pursued through random modality dropout
\cite{cite:bevfusion_mit, cite:unibev} and through fusion stages
that gracefully degrade when a stream is missing or impaired
\cite{cite:metabev, cite:cafuser, cite:resilient}. The works most
closely related to ours target adverse weather or sensor faults at
the architectural level. CrossFuser~\cite{cite:crossfuser} relies on
a domain-shift module trained on weather-perturbed inputs.
PolicyFuser \cite{cite:policyfuser} abandons feature-level fusion
altogether and reconciles per-modality policies.
MaskFuser~\cite{cite:maskfuser} trains a unified token space with
cross-modal masked auto-encoding. CAFuser~\cite{cite:cafuser}
classifies the environmental condition into a coarse, scene-level
token that biases the fusion stack. These methods improve robustness,
but their reliability cues are either modality-level, token-reconstruction
based or scene-level. They do not explicitly represent the physical
severity of each corrupted cell. Hence, they can decide that a stream or
scene is unreliable, but cannot provide the planner with a dense spatial
selector that suppresses only the damaged regions while preserving the
unaffected regions of the same modality. Each of these methods therefore
operates at a granularity coarser than the per-cell signal that
asymmetric corruption demands. None of them uses a calibrated physical
quantity as the gating input, and none exposes a per-cell reliability map
that the planner can consume directly as a structural feature.

\subsection{Positioning of the Proposed Framework}
VG-SAF intersects all three threads. Its positioning can be summarized
by the same three difficulties identified in the Introduction. To obtain
dense reliability supervision, it pairs the variance head with a
physically grounded augmentor that emits a continuous corruption mask. To
avoid an unstable residual-only uncertainty proxy, it uses cross-branch
dense distillation in log space, which produces a monotone
severity-to-reliability map without requiring per-mode labels. To make the
variance actionable for planning, it converts the calibrated map into a
hybrid attention composed of a within-modality local gate and a
cross-modal trust softmax. The distinguishing feature of VG-SAF is thus
not a post-hoc uncertainty diagnostic, but the integration of
mask-calibrated dense reliability, physical mask supervision and
local-plus-trust attention within a single end-to-end driving stack and
coherent training pipeline.

\section{Methodology}\label{sec:method}

\subsection{Overview}\label{subsec:overview}
\begin{figure*}[!t]
\centering
\includegraphics[width=\textwidth]{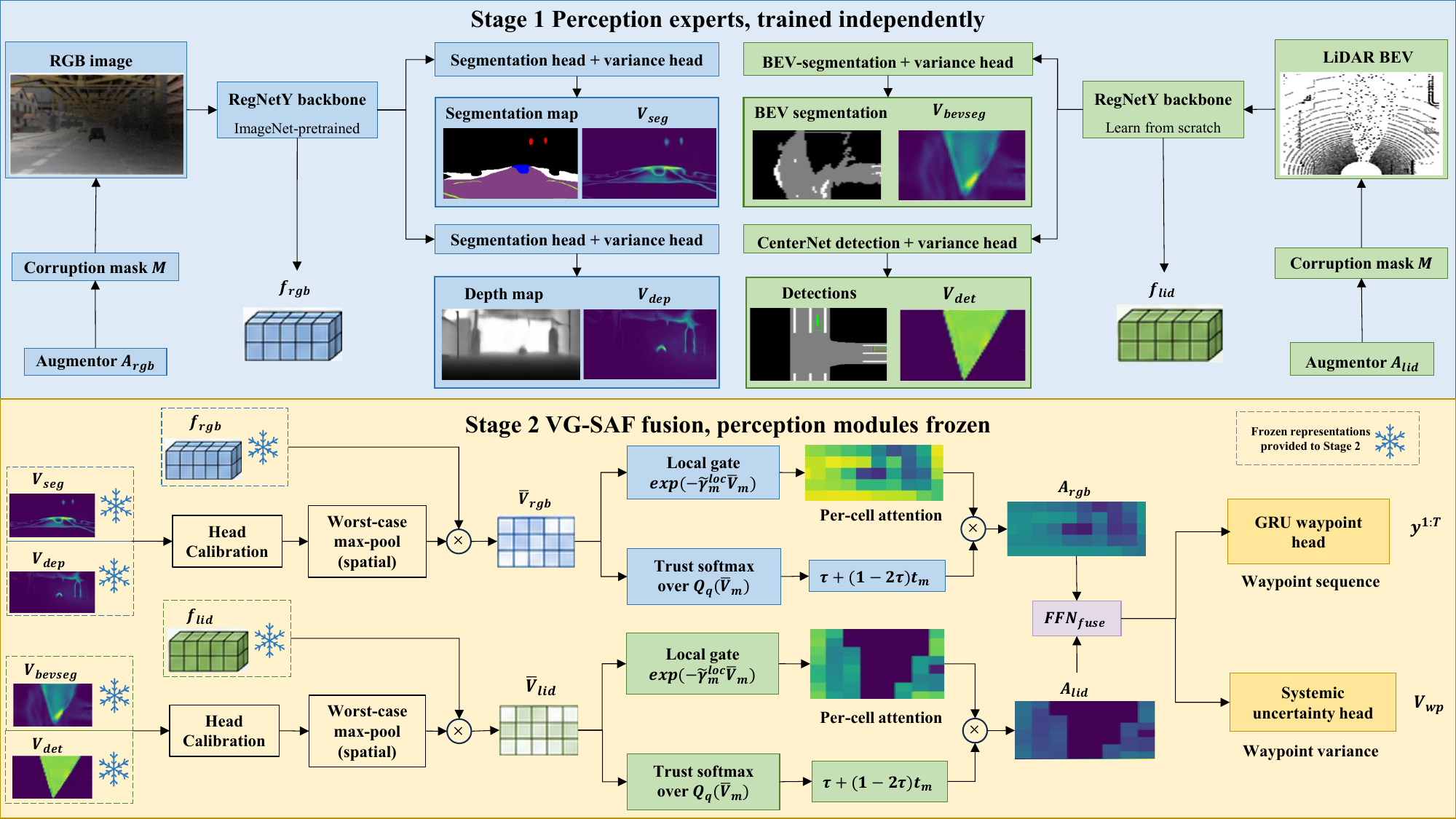}
\caption{Two-stage training workflow of VG-SAF. Stage~1 (top) trains
two independent perception experts under the per-modality augmentors
$\mathcal{A}_{\mathrm{rgb}}$ and $\mathcal{A}_{\mathrm{lid}}$ and
the perception loss $\mathcal{L}_{\mathrm{percep}}$. Stage~2 (bottom)
freezes the experts and learns the fusion stage, where dense reliability
maps are aggregated into $\bar V_m$ and converted into a local gate and
a trust softmax. Their product forms $\mathbf{A}_m$, which gates each
modality before waypoint decoding.}
\label{fig:workflow}
\end{figure*}

VG-SAF is organized into two cascaded stages, both supervised by a
single physical signal: the continuous per-pixel corruption mask
$\mathbf{M}\!\in\![0,1]^{H\times W}$ emitted by the augmentor
(Fig.~\ref{fig:workflow}). Stage~1 trains two modality-specific
\emph{uncertainty-aware perception experts}
(Section~\ref{subsec:perception}). Stage~2 freezes the experts and
trains the fusion components (Section~\ref{subsec:vgsaf}). The design
therefore follows a clear information path: the augmentor records where
physical damage is injected, the perception experts convert this record
into calibrated dense variance, and the fusion stage uses the variance
to suppress spatially unreliable cells before waypoint decoding. A
complete description of the network architecture, the augmentor catalog
and the curriculum schedules is provided in the accompanying
supplementary material.

\subsection{Network Architecture and Inputs}\label{subsec:arch}
The camera input is the RGB frame
$\mathbf{x}_{\mathrm{rgb}}\!\in\!\mathbb{R}^{3\times 160\times 704}$.
The LiDAR input is a two-channel BEV histogram
$\mathbf{x}_{\mathrm{lid}}\!\in\!\mathbb{R}^{2\times 256\times 256}$
at $8$ pixels per meter. The third input is the navigation goal
$\mathbf{p}_{t}\!\in\!\mathbb{R}^{2}$ in the ego-vehicle frame. Both
modality experts adopt the RegNetY backbone family
\cite{cite:regnety}. The camera backbone is initialized from
ImageNet weights. The LiDAR backbone reuses the architecture but is
trained from scratch with a freshly initialized stem. The camera
expert produces a per-pixel semantic segmentation map with variance
$V^{\mathrm{seg}}$ and a depth map with variance $V^{\mathrm{dep}}$.
The LiDAR expert produces a BEV segmentation map with variance
$V^{\mathrm{bev}}$ and a CenterNet detection head
\cite{cite:centernet} with a per-pixel detection variance
$V^{\mathrm{det}}$. The fusion stage adds four learnable scalars, a
fusion projection $\mathrm{FFN}_{\mathrm{fuse}}$, an autoregressive
GRU waypoint head and a systemic uncertainty head. Layer widths,
head dimensions and the GRU recurrence schedule are detailed in
Supplementary Section~S2, including the exact spatial resolutions used
by the fusion head.

\subsection{Augmentor and Corruption Mask}\label{subsec:augment}
Each modality is paired with an online augmentor
$\mathcal{A}_m$. Given a clean input, the augmentor samples one
failure mode from a phase-dependent distribution. The output is the
corrupted input plus three companions: a dense corruption mask
$\mathbf{M}\!\in\![0,1]^{H\times W}$, a sample-level corruption
severity $K\!=\!100\,\mathbb{E}[\mathbf{M}]\%$, and a mode identifier. A value
of zero in $\mathbf{M}$ denotes an untouched pixel; a value of one
denotes a fully destroyed pixel; intermediate values denote partial
damage. The mask is \emph{global} when its value is uniform across
the frame (e.g., sensor saturation, calibration drift) and
\emph{local} when its support is geometrically bounded (e.g., mud on
the lens, an angular LiDAR wedge). The continuous per-pixel form of
$\mathbf{M}$ provides dense supervision for the reliability head without additional annotation.

The camera augmentor contains eight optical failure modes. The
\texttt{signal\_drop} mode creates blackout, whiteout or snow-like
saturation; \texttt{local\_noise} injects low-light noise patches;
\texttt{local\_exposure\_pulse} models local flare and bloom;
\texttt{local\_occlusion} emulates mud or debris on the lens;
\texttt{night\_lowlight} applies global gain and gamma loss;
\texttt{motion\_blur} applies directional convolution from vibration;
\texttt{ghosting} creates shifted translucent copies; and
\texttt{color\_shift} models white-balance or color-calibration drift.
The LiDAR augmentor contains five geometric failure modes:
\texttt{signal\_drop} simulates frame-wide jamming or spurious returns;
\texttt{range\_dropout} removes returns in a distance-dependent way;
\texttt{frustum\_occlusion} blocks an angular wedge; \texttt{local\_speckle}
injects dust-like false returns; and \texttt{feature\_noise} adds
channel-wise perturbations to active voxels. Together, these modes cover
both global sensor-side degradation and local field-of-view damage, which
is the asymmetric setting where dense masks provide more information than
a modality-level missing-sensor flag. The augmentor should not be read as
a complete model of all real sensor failures; rather, it provides
physically motivated supervision for the reliability field. To reduce the
risk of evaluating only the exact training perturbations, the closed-loop
routes, random seeds and sampled severities used for evaluation are held
out from training, and the attention ablation further uses generic
uniform and elliptical masks that are not tied to a particular named
fault mode. The training ranges, real-world analogues, severity-extrapolation protocol,
and example renderings are given in Supplementary Section~S3.

\subsection{Uncertainty-Aware Perception Experts}\label{subsec:perception}
Each task head shares a decoder body with a reliability head. The
segmentation and detection scales use
$V\!=\!\mathrm{sp}(\rho)\!+\!10^{-2}$, whereas the depth head predicts
$\sigma\!=\!0.05+\mathrm{sp}(\rho)$ and uses
$V^{\mathrm{dep}}\!=\!\sigma^2$ (Supplementary Section~S2-C). We write
the common loss-attenuation form as
\begin{equation}
\mathcal{L}_{\mathrm{het}}(\hat y,y;V)=a_k\!\left(
\frac{\ell_k(\hat y,y)}{V}+\log V\right),
\label{eq:gnll}
\end{equation}
where $a_k\!=\!\tfrac12$ and $\ell_k$ is squared error for the Gaussian
depth model. For segmentation and detection heatmaps, Eq.~\eqref{eq:gnll}
is used as an uncertainty-weighted cross-entropy surrogate with a
constant task scaling; the detection regression branches use the
Laplace NLL \cite{cite:kendall_gal}. For the generic attenuated form, the
stationary scale is proportional to the local supervised loss, so the
head can self-organize as a predictor of loss magnitude. Physical
corruption, however, is not identical to supervised loss: a severely
damaged observation may contain insufficient evidence for reliable
direct supervision, while an easy clean cell may still have a small
loss. The mask-based terms below therefore align the predicted scale
explicitly with physical corruption severity.

From Phase~2 onward, the encoder ingests a concatenated batch
$[\mathrm{clean}|\mathrm{noisy}]$ and emits clean and noisy halves of
every prediction. Let $(\hat y_c,V_c)$ and $(\hat y_n,V_n)$ denote the
corresponding task predictions and variances, and let
$\mathbf{W}=\mathbf{1}-\mathbf{M}$ be the trustworthiness mask. The
ordinary task supervision is retained on the clean branch and weighted
continuously on the noisy branch according to the remaining observation
reliability. This mask-weighted supervised
loss is
\begin{equation}
\mathcal{L}_{\mathrm{sup}}
=\mathcal{L}_{\mathrm{het}}(\hat y_c,y;V_c)
+\lambda_{\mathrm{sup}}\,\mathbf{W}\odot
\mathcal{L}_{\mathrm{het}}(\hat y_n,y;V_n).
\label{eq:supervision}
\end{equation}
For severely corrupted cells, the noisy observation may no longer contain
sufficient evidence for reliable direct supervision. The weight
$\mathbf{W}=\mathbf{1}-\mathbf{M}$ therefore preserves full supervision on
clean cells and progressively reduces it as corruption severity increases. We further add a
confidence-aware consistency term that pulls the noisy branch back to
the clean branch only where the augmentor leaves the input reliable:
classification heads use
$\mathbf{W}\odot\mathrm{KL}(\mathbf{P}_n\|\mathrm{sg}(\mathbf{P}_c))$,
where $\mathbf{P}=\mathrm{softmax}(\mathbf{z})$, while regression heads
use $\mathbf{W}\odot(\hat y_n-\mathrm{sg}(\hat y_c))^2$. The
stop-gradient operator prevents the noisy branch from reshaping the
clean reference during dual-branch calibration.

The novel calibration term is what we refer to as
\emph{cross-branch dense distillation}. Its novelty lies in using the
clean branch as a self-reference while using the augmentor mask as a
dense physical severity label. The term therefore avoids manual
uncertainty annotation and prevents the noisy branch from learning a
variance map driven only by task residuals. Let $V_c$ and $V_n$ be the
clean and noisy variance predictions. We define a per-pixel target on
the noisy log-variance that is linear in the local corruption severity:
\begin{equation}
\log V_{\mathrm{tgt}} = \mathrm{sg}(\log V_c) + \alpha\,\mathbf{M},
\label{eq:caltgt}
\end{equation}
\begin{equation}
\mathcal{L}_{\mathrm{cal}} = \mathrm{SmoothL1}(\log V_n, \log V_{\mathrm{tgt}}),
\label{eq:calloss}
\end{equation}
with $\mathrm{sg}(\cdot)$ the stop-gradient operator and $\alpha\!>\!0$
a head-specific scaling constant. The multiplicative form
$V_{\mathrm{tgt}}\!=\!V_c\exp(\alpha\mathbf{M})$ is equivalent. On
clean pixels the noisy variance is distilled to match the clean
variance. On fully destroyed pixels the noisy variance is targeted
at $e^{\alpha}$ times the clean variance. The cross-entropy heads
use $\alpha\!=\!3$. The focal-loss CenterNet head uses
$\alpha\!=\!4.6$, which compensates for the smaller baseline
variance produced by its focal-loss supervisor \cite{cite:focal}.

The total perception loss at epoch $e$ is
\begin{equation}
\begin{aligned}
\mathcal{L}_{\mathrm{percep}}^{m}(e)
&= \sum_{k\in\mathcal{T}_m}\mathcal{L}^{(k)}_{\mathrm{sup}} \\
&\quad + \lambda_{\mathrm{con}}(e)\sum_{k\in\mathcal{T}_m}
\mathcal{L}^{(k)}_{\mathrm{con}}
+ \lambda_{\mathrm{cal}}(e)\sum_{k\in\mathcal{T}_m}
\mathcal{L}^{(k)}_{\mathrm{cal}},
\end{aligned}
\label{eq:totalpercep}
\end{equation}
where $\mathcal{T}_{\mathrm{rgb}}\!=\!\{\mathrm{seg},\mathrm{dep}\}$ and
$\mathcal{T}_{\mathrm{lid}}\!=\!\{\mathrm{bev},\mathrm{det}\}$. The three
terms have distinct roles: $\mathcal{L}_{\mathrm{sup}}$ keeps the clean
branch accurate and trains the noisy branch only on reliable cells;
$\mathcal{L}_{\mathrm{con}}$ preserves prediction consistency on cells
that remain visually or geometrically valid; and $\mathcal{L}_{\mathrm{cal}}$
turns the automatically emitted corruption mask into dense reliability
supervision. The curriculum weights $\lambda_{\mathrm{con}}(e)$ and
$\lambda_{\mathrm{cal}}(e)$ are activated gradually to avoid forcing
prediction consistency and variance inflation at full strength from the
same early mini-batches.

The four-phase curriculum builds the variance head incrementally.
Phase~1 trains a single-branch encoder with all calibration weights
at zero, producing a stable backbone. Phase~2 introduces the
dual-branch regime and ramps the consistency and calibration weights on
\emph{staggered} schedules to avoid the gradient conflict between
``pull the noisy branch back to the clean prediction'' and
``inflate $\log V$ on the same cells''. Phase~3 consolidates the
joint solution under a uniform mode mix at full weights. Phase~4
shifts the mode distribution toward the physically severe failures, so
the final variance head is calibrated on the corruptions that most affect
closed-loop driving. During Phases~2--4, all BatchNorm layers keep their
running statistics frozen while their affine parameters remain trainable.
This simple freeze is important because mixed clean--noisy batches would
otherwise make the running statistics represent neither distribution; at
inference the mismatch can drive variance pre-activations toward the
softplus floor and produce a collapsed or unresponsive uncertainty head. The
complete curriculum and BatchNorm implementation are documented in
Supplementary Section~S6.

\subsection{Variance-Guided Spatial Attention Fusion}\label{subsec:vgsaf}

The fusion stage converts the calibrated reliability field into spatial
attention before the waypoint head receives the features. The term
\emph{spatial} is used deliberately: the attention is not a single
sample-level confidence score but a per-cell multiplicative field on the
RGB and LiDAR feature grids. We first normalize the task-specific
variance maps so their magnitudes are comparable, then pool them onto
the fusion resolution, and finally combine two reliability mechanisms: a
local gate that suppresses unreliable cells within a modality and a
trust softmax that redistributes evidence across modalities.

\subsubsection{Variance Normalization and Pooling}
The four per-task variance maps live on different magnitude scales.
We normalize each by a clean-data empirical mean
$n^{\mathrm{seg,rgb}},n^{\mathrm{dep}},n^{\mathrm{seg,lid}}$, and $n^{\mathrm{det}}$
so that each task-specific variance channel has a unit-mean baseline
before summation:
\begin{equation}
V_{\mathrm{rgb}} = \frac{V^{\mathrm{seg}}}{n^{\mathrm{seg,rgb}}}
 + \frac{V^{\mathrm{dep}}}{n^{\mathrm{dep}}},\;
V_{\mathrm{lid}} = \frac{V^{\mathrm{bev}}}{n^{\mathrm{seg,lid}}}
 + \frac{V^{\mathrm{det}}}{n^{\mathrm{det}}}.
\label{eq:vsum}
\end{equation}
The normalizers are computed once on a clean validation split and then
frozen, so the fusion stage receives comparable reliability scales
without learning a static offset between task heads. This design keeps
large values interpretable as corruption-induced uncertainty rather than
as a consequence of task-dependent loss units. The two
per-modality variance maps are then resampled onto the corresponding
feature grid by worst-case max-pooling:
\begin{equation}
\bar V_m = \mathrm{maxpool}_{h_m\times w_m}(V_m),
\quad m\!\in\!\{\mathrm{rgb},\mathrm{lid}\}.
\label{eq:vpool}
\end{equation}
A single cell of high uncertainty within a pooling window is
sufficient evidence to down-weight the corresponding pooled feature
stack. Max-pooling is therefore the natural aggregator for the
gating task within our framework, and the worst-case operator is
preferred over averaging for safety-critical gating decisions.

\subsubsection{Hybrid Variance-Guided Attention}
The fusion stage uses four unconstrained learnable scalars,
$\gamma^{\mathrm{loc}}_m$ and $\gamma^{\mathrm{tru}}_m$ for
$m\!\in\!\{\mathrm{rgb},\mathrm{lid}\}$. They are mapped to positive
gains $\eta^{\mathrm{loc}}_m=\mathrm{sp}(\gamma^{\mathrm{loc}}_m)+10^{-4}$
and $\eta^{\mathrm{tru}}_m=\mathrm{sp}(\gamma^{\mathrm{tru}}_m)+10^{-4}$
before use. Positivity enforces the intended monotonic relation: higher
variance can only reduce local attention or modality trust, and can never
reward an unreliable region. The within-modality spatial gate is
\begin{equation}
\mathbf{A}^{\mathrm{loc}}_m = \exp(-\eta^{\mathrm{loc}}_m\,\bar V_m)
 \in (0,1].
\label{eq:Aloc}
\end{equation}
The cross-modal trust scalar is a softmax over the negatively-scaled
lower-quartile summaries $s^V_m = Q_{0.25}(\bar V_m)$ of each
modality's variance map:
\begin{equation}
[t_{\mathrm{rgb}}, t_{\mathrm{lid}}] =
 \mathrm{softmax}\!\Bigl(\bigl[-\eta^{\mathrm{tru}}_{\mathrm{rgb}} s^V_{\mathrm{rgb}},
                              -\eta^{\mathrm{tru}}_{\mathrm{lid}} s^V_{\mathrm{lid}}\bigr]\Bigr).
\label{eq:trust}
\end{equation}
The $25$th percentile aggregator asks whether the modality has
enough reliable cells to be worth trusting, and is insensitive to
isolated noisy cells that the local gate already handles. A trust
floor $\tilde t_m\!=\!\tau\!+\!(1\!-\!2\tau) t_m$ with
$\tau\!=\!0.15$ prevents complete over-suppression of either
modality. The floor is a safety guard rather than a performance trick:
when one stream is globally degraded, it may still contain local cues,
such as lane boundaries or traffic-light evidence, that should not be
irreversibly discarded. The combined per-cell attention is
\begin{equation}
\mathbf{A}_m = \tilde t_m\,\mathbf{A}^{\mathrm{loc}}_m,
\quad
\tilde{\mathbf{f}}_m = \mathbf{A}_m \odot \mathbf{f}_m.
\label{eq:Atotal}
\end{equation}
The two factors are deliberately driven by independent gains. The
local gate prefers a large $\eta^{\mathrm{loc}}$ for a sharp
per-cell attenuation. The trust softmax prefers a small
$\eta^{\mathrm{tru}}$ for a smooth cross-modal split. A shared
gain would force a compromise between these two regimes. The
explicit decoupling resolves the tension at no additional parameter
cost and with negligible architectural overhead or additional sensor inputs.

\subsubsection{Fusion, Waypoints and Systemic Uncertainty}
The two gated feature maps are global-average-pooled, concatenated
and projected back through $\mathrm{FFN}_{\mathrm{fuse}}$ to a fused
representation $\mathbf{z}_{\mathrm{fuse}}\!\in\!\mathbb{R}^C$
(Supplementary Section~S2-D). The waypoint sequence
$\hat{\mathbf{y}}_{1:T}$ is decoded autoregressively by a GRU
\cite{cite:gru, cite:transfuser}. In parallel, a systemic
uncertainty head emits a positive Laplace scale $V_{\mathrm{wp}}$ from the fused state
and the two worst-case modality summaries:
\begin{equation}
\begin{aligned}
V_{\mathrm{wp}}=\mathrm{sp}\!\bigl(\mathrm{MLP}(&[\mathbf{z}_{\mathrm{fuse}};
\max_{u,v}\bar V_{\mathrm{rgb}}(u,v);\\
&\max_{u,v}\bar V_{\mathrm{lid}}(u,v)])\bigr)+\varepsilon.
\end{aligned}
\label{eq:Vwp}
\end{equation}
This design exposes severe local sensor failure to the waypoint head
even after global feature pooling. The planning objective is the two-dimensional Laplace negative
log-likelihood, up to constants independent of the model,
\begin{equation}
\mathcal{L}_{\mathrm{wp}} = \frac{1}{T}\sum_{t=1}^{T}
\left(
\frac{\|\hat{\mathbf{y}}_t-\mathbf{y}_t\|_1}{V_{\mathrm{wp}}}
+ 2\log V_{\mathrm{wp}}
\right),
\label{eq:wpnll}
\end{equation}
where $V_{\mathrm{wp}}$ denotes a trajectory-level Laplace scale, not a
variance, shared across the $T$ waypoint steps and the two waypoint
coordinates. The factor $2\log V_{\mathrm{wp}}$ accounts for the two
independent coordinates. The ground-truth waypoints are never altered by
the augmentor. When the $\ell_1$ residual is large under severe
corruption, Eq.~\eqref{eq:wpnll} increases the predicted scale rather
than forcing the GRU to fit unreliable evidence. Large
$V_{\mathrm{wp}}$ therefore provides a safety signal for severe,
combined, or severity-extrapolation faults. A light trust-balance regularizer keeps the no-fault trust split near the
$0.5{:}0.5$ prior:
\begin{equation}
\mathcal{L}_{\mathrm{tb}}
=\frac{1}{|\mathcal{I}_c|}
\sum_{i\in\mathcal{I}_c}
\left(t_{\mathrm{rgb}}^{(i)}-\tfrac{1}{2}\right)^2,
\label{eq:trustbalance}
\end{equation}
where $\mathcal{I}_c$ is the set of clean mini-batch samples. The
regularizer is applied only on no-fault inputs, so it prevents silent drift
toward a single dominant modality without weakening the intended
corruption-dependent trust reweighting. The full fusion-stage curriculum
is documented in Supplementary Section~S6-C.

\section{Experiments and Results}\label{sec:experiments}

\subsection{Experimental Setup}\label{subsec:setup}
All experiments use CARLA version $0.9.10$ with raw
$480\!\times\!800$ front-camera images and a $64$-line LiDAR. The
camera frames are resized to $160\!\times\!704$ before entering the
network, preserving the horizontal field of view used by the planner. Training data
is collected from the privileged autopilot agent across the eight
training towns and consists of approximately $300\mathrm{k}$ frames at
$2\,\mathrm{Hz}$. We evaluate on Longest6 \cite{cite:fusion_transformer},
a curated set of $36$ routes of average length $1.5\,\mathrm{km}$
across six towns. Each route is replayed under three corruption
regimes: camera corrupted, LiDAR corrupted, and both modalities
corrupted. For every route we independently sample one camera mode
and one LiDAR mode from the augmentor's catalog and apply the selected
augmentations to every frame. To avoid a trivial closed loop between
training and testing, Longest6 routes are never used for training, and
the evaluation uses held-out random seeds and route-level persistent
corruptions. In addition to held-out in-range severities, selected tests
extrapolate $K$ beyond the training ranges to assess robustness to unseen
fault intensity. The ablation in
Section~\ref{subsec:ablation} further tests generic uniform and
elliptical masks, which are not exact copies of any training mode.
Reported numbers are averaged over three independent runs of the full
route set per regime, with seeds drawn from a fixed pool to ensure
reproducibility across the three corruption regimes.

We use \emph{robustness} to mean preservation of closed-loop driving
performance under these sensor-degradation regimes relative to the same
backbone family without variance-guided gating. Accordingly, we adopt
the three primary CARLA evaluation metrics. Route
completion $\mathrm{RC}\!\in\![0,100]\%$ is the route distance
completed before termination. Infraction score
$\mathrm{IS}\!\in\![0,1]$ is a multiplier starting at unity that
reduces multiplicatively per infraction. Driving score
$\mathrm{DS}\!=\!\mathrm{RC}\!\cdot\!\mathrm{IS}$ is the official
ranking metric. Higher is better for all three. Stage~1 trains for
$140$ epochs and Stage~2 for $100$ epochs on two NVIDIA RTX~6000 Ada
GPUs. Inference runs at $34\,\mathrm{ms}$ per frame. Additional compute,
scheduling and reproducibility details are documented in Supplementary
Section~S8.

\subsection{Closed-Loop Longest6 Comparison under Sensor Corruption}\label{subsec:leaderboard}

We compare against four baselines that span the principal design
choices in end-to-end multimodal driving. The \textbf{Image-only}
baseline corresponds to the Latent TransFuser variant of
\cite{cite:fusion_transformer}. The \textbf{LiDAR-only} baseline has
no access to camera information, and therefore no access to
traffic-light phase. \textbf{Equal-Weight Fusion} concatenates the
two latent feature maps with fixed equal weights, matching the Late
Fusion variant of \cite{cite:fusion_transformer} \cite{cite:transfuser}.
\textbf{TransFuser}~\cite{cite:fusion_transformer}\cite{cite:transfuser} interconnects the
two backbones through multi-scale cross-attention. All baselines
share the same encoder backbones, the same training data and the same
waypoint head as our VG-SAF implementation. This controlled setting
aligns the non-fusion components and focuses the comparison on the
fusion strategy. Robust fusion methods such as CrossFuser, MaskFuser,
PolicyFuser and CAFuser are discussed in Section~\ref{sec:related}; a
fully reproduced numerical comparison with them would require matching
their training recipes, perception heads and closed-loop evaluation
interfaces. We therefore use the controlled TransFuser-family comparison
as the primary evidence for the proposed fusion mechanism, and report
component ablations below to isolate the effect of variance-guided local
and trust attention.

\begin{table}[!t]
\caption{Closed-loop Longest6 results under three sensor-corruption
regimes (mean$\,\pm\,$std over three runs). Metrics are averaged at the
route/run level; therefore, the product of the displayed mean RC and mean
IS need not equal the displayed mean DS.}
\label{tab:longest6_corruption}
\centering
\footnotesize
\begin{tabular}{lccc}
\toprule
Model & $\mathrm{DS}\!\uparrow$ & $\mathrm{RC}\!\uparrow$ & $\mathrm{IS}\!\uparrow$ \\
\midrule
\multicolumn{4}{l}{\textit{Camera corrupted (LiDAR clean)}} \\
Image-only            & $18.4\pm 6.1$ & $73.2\pm 8.7$ & $0.25\pm 0.06$ \\
Equal-Weight Fusion   & $14.7\pm 5.8$ & $68.5\pm 9.4$ & $0.22\pm 0.05$ \\
TransFuser            & $32.1\pm 7.3$ & $84.6\pm 6.2$ & $0.40\pm 0.07$ \\
\textbf{VG-SAF (Ours)}
                      & $\mathbf{42.6\pm 5.9}$
                      & $\mathbf{91.2\pm 4.1}$
                      & $\mathbf{0.51\pm 0.06}$ \\
\midrule
\multicolumn{4}{l}{\textit{LiDAR corrupted (camera clean)}} \\
LiDAR-only            & $16.8\pm 6.5$ & $67.4\pm 9.1$ & $0.23\pm 0.06$ \\
Equal-Weight Fusion   & $16.2\pm 6.0$ & $71.8\pm 8.5$ & $0.24\pm 0.05$ \\
TransFuser            & $36.4\pm 6.8$ & $88.7\pm 5.4$ & $0.44\pm 0.07$ \\
\textbf{VG-SAF (Ours)}
                      & $\mathbf{44.9\pm 5.4}$
                      & $\mathbf{92.5\pm 3.6}$
                      & $\mathbf{0.53\pm 0.05}$ \\
\midrule
\multicolumn{4}{l}{\textit{Both modalities corrupted}} \\
Equal-Weight Fusion   & $10.8\pm 5.2$ & $62.3\pm 10.1$& $0.18\pm 0.05$ \\
TransFuser            & $24.5\pm 7.0$ & $77.4\pm 7.9$ & $0.34\pm 0.07$ \\
\textbf{VG-SAF (Ours)}
                      & $\mathbf{38.2\pm 6.1}$
                      & $\mathbf{86.8\pm 4.8}$
                      & $\mathbf{0.47\pm 0.06}$ \\
\bottomrule
\end{tabular}
\end{table}

Table~\ref{tab:longest6_corruption} reports the three primary
metrics. VG-SAF achieves the highest $\mathrm{DS}$ in every regime
and exceeds TransFuser by $10.5$, $8.5$ and $13.7$ points under
camera, LiDAR and both-modality corruption respectively. Three
patterns emerge. The Equal-Weight Fusion baseline is consistently
the weakest, because the impaired stream is averaged with the clean
stream and the corrupted bias propagates into the fused
representation. The single-modality baselines are intermediate but
structurally limited; the LiDAR-only baseline frequently violates red
lights even on routes where the geometry is recoverable. TransFuser
is more robust under LiDAR corruption than under camera corruption,
because its cross-attention features rely more heavily on the camera
in urban scenarios; however, its fusion weights are not conditioned
on input quality, so the corrupted modality continues to influence
the fused features in proportion to its trained attention weight.
VG-SAF preserves the same intra-regime ordering as TransFuser but
narrows the cross-regime degradation from $4.3$ to $2.3$
$\mathrm{DS}$ points, evidence that the trust softmax absorbs the
asymmetric reliance on the camera by down-weighting whichever stream
becomes locally noisier.

\subsection{Ablation Studies}\label{subsec:ablation}

We isolate the contribution of each attention factor through a
controlled corruption sweep. The sweep applies the unified corruption severity
$K\!=\!100\,\mathbb{E}[\mathbf{M}]\%\in[0,100]\%$ to either modality independently or to both
together, using two complementary schemes (uniform random masking
and elliptical occlusions) whose outputs are pooled at evaluation
time. Every forward pass is repeated under four attention
configurations $A_m\!\in\!\{1, A^{\mathrm{loc}}_m, t_m, t_m
A^{\mathrm{loc}}_m\}$, denoted \emph{none}, \emph{local},
\emph{trust} and \emph{full} in the remainder of this section.

\begin{figure}[!t]
\centering
\includegraphics[width=\columnwidth]{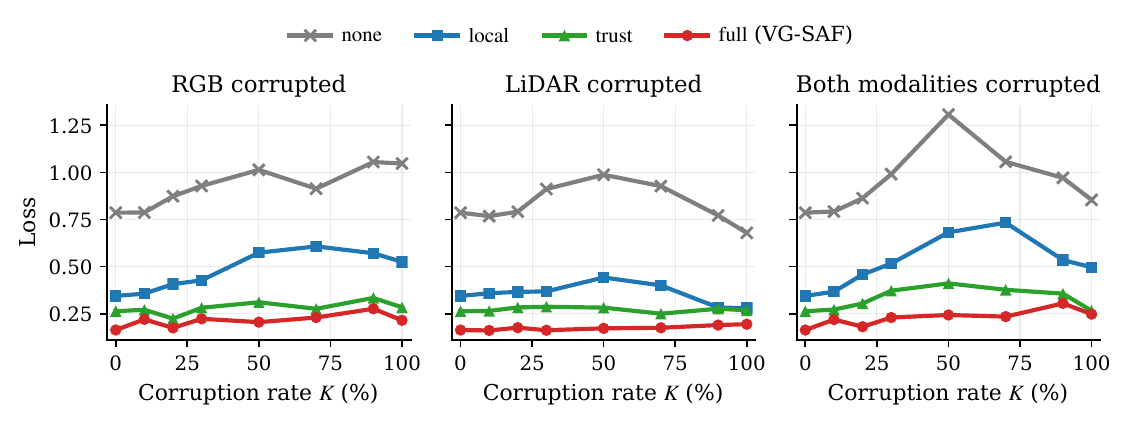}
\caption{Median waypoint $L_1$ error (m) versus corruption severity
$K$ under four attention configurations and three corruption scenarios.
For each setting, errors are first averaged over the four future
waypoints of each frame and the median is then taken over all evaluated
frames, routes, and runs. Lower is better.}
\label{fig:loss_vs_K}
\end{figure}

\begin{figure}[!t]
\centering
\includegraphics[width=\columnwidth]{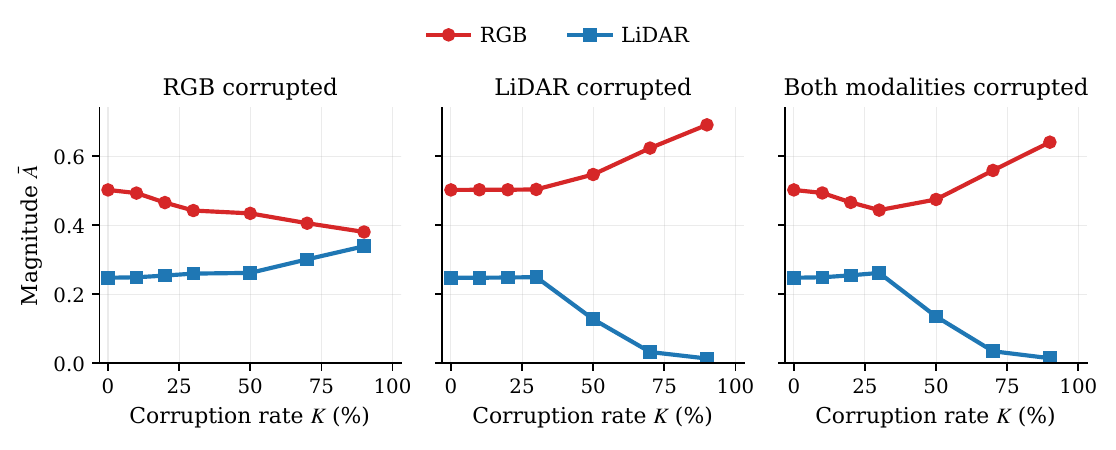}
\caption{Fused per-modality attention magnitude $\bar A_m$ versus
corruption severity $K$ under three scenarios. Here, $\bar A_m$ is the
spatial mean of $\mathbf A_m$, averaged over all evaluated frames,
routes, and runs. The impaired modality is monotonically down-weighted
while the intact modality absorbs the displaced mass. Data stop at
$K\!=\!90\%$; the $100\%$ tick is retained only for scale.}
\label{fig:attn_vs_K}
\end{figure}

\begin{table}[!t]
\caption{Per-mode reliability response in the upper severity tertile of
each mode. $K$ is the mean mask severity; $\bar V$ is the pooled
per-modality reliability scale averaged over all qualifying frames from
36 Longest6 routes and three runs; $\bar V/\bar V_{\mathrm{clean}}$ is
the clean-normalized ratio; and $\Delta t$ is the corrupted modality's
trust shift relative to clean operation. Values above the training ranges
in Tables~S1--S2 are deliberate severity-extrapolation tests.}
\label{tab:per_mode_v}
\centering
\footnotesize
\begin{tabular}{lcccc}
\toprule
Modality/Mode & $K$ (\%) & $\bar V$ & $\bar V/\bar V_{\mathrm{clean}}$ & $\Delta t$ \\
\midrule
\multicolumn{5}{l}{\textbf{\textit{Camera}}} \\
signal\_drop          & 100 & 37.26 & 37.5$\times$ & $-0.11$ \\
local\_noise          & 45 &  5.77 &  5.8$\times$ & $-0.01$ \\
local\_exposure\_pulse& 70 & 22.67 & 22.8$\times$ & $-0.09$ \\
local\_occlusion      & 59 & 11.75 & 11.8$\times$ & $-0.02$ \\
night\_lowlight       & 60 &  7.03 &  7.1$\times$ & $-0.01$ \\
motion\_blur          & 91 & 28.93 & 29.1$\times$ & $-0.07$ \\
ghosting              & 31 &  3.57 &  3.6$\times$ & $-0.01$ \\
color\_shift          & 11 &  1.08 &  1.1$\times$ & $\phantom{-}0.00$ \\
\midrule
\multicolumn{5}{l}{\textbf{\textit{LiDAR}}} \\
signal\_drop          & 100 &105.25 & 57.9$\times$ & $-0.47$ \\
range\_dropout        & 52 & 11.31 &  6.2$\times$ & $-0.10$ \\
frustum\_occlusion    & 64 & 68.58 & 37.7$\times$ & $-0.33$ \\
local\_speckle        & 18 &  5.87 &  3.2$\times$ & $-0.02$ \\
feature\_noise        & 54 & 10.33 &  5.7$\times$ & $-0.16$ \\
\bottomrule
\end{tabular}
\end{table}

\begin{figure*}[!t]
\centering
\includegraphics[width=0.9\textwidth]{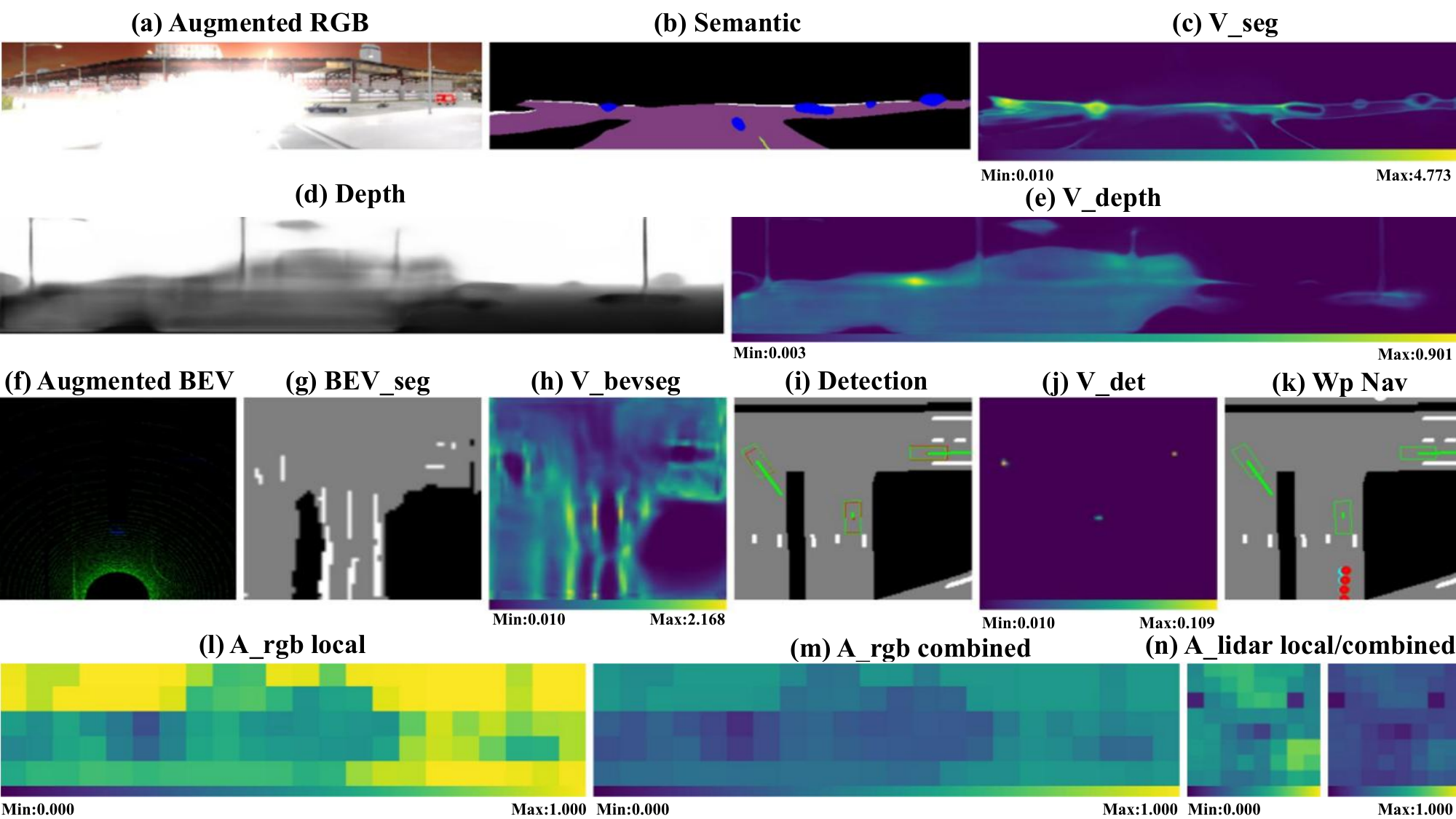}
\caption{Camera \textit{local\_exposure\_pulse} fault at $K=48\%$; LiDAR clean. The dense reliability response raises the RGB-side uncertainty and suppresses unreliable local features while LiDAR remains trusted.}
\label{fig:qualitative_rgb}
\end{figure*}

\begin{figure*}[!t]
\centering
\includegraphics[width=0.9\textwidth]{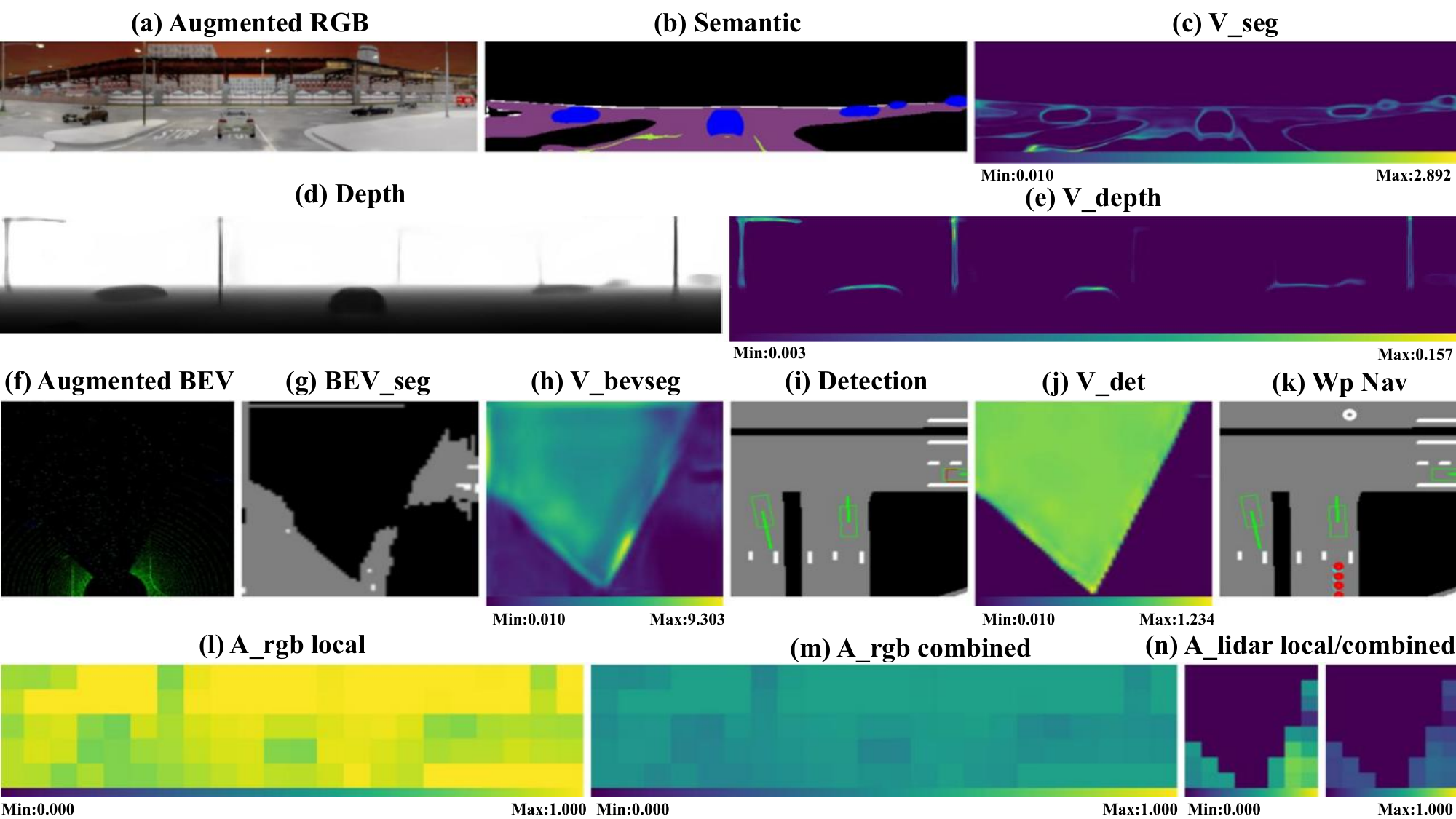}
\caption{LiDAR \textit{frustum\_occlusion} fault at $K=59\%$; camera clean. The BEV reliability scale peaks inside the missing wedge, down-weighting corrupted LiDAR cells and preserving the camera-supported trajectory.}
\label{fig:qualitative_lidar}
\end{figure*}

Fig.~\ref{fig:loss_vs_K} shows that the four curves are well
separated at every $K$. The attention-free baseline reaches a peak
$L_1$ near $1.3\,\mathrm{m}$ in the both-corrupted scenario at
$K\!\approx\!50\%$; the \emph{trust} and \emph{full} configurations
remain near $0.4\,\mathrm{m}$ and $0.3\,\mathrm{m}$ respectively at
the same operating point. The trust-over-local gap is roughly $40\%$
of the local loss, while the full-over-trust gap is roughly $30\%$ of
the trust loss. The trust softmax therefore carries the majority of
the robustness gain under heavy corruption, while the local gate
contributes a complementary correction on the residual
within-modality damage that survives the trust re-weighting stage of
the fusion pipeline.

Fig.~\ref{fig:attn_vs_K} confirms the predicted redistribution. When
only the LiDAR is corrupted, $\bar A_{\mathrm{lid}}$ decreases
monotonically from $0.25$ at $K\!=\!0\%$ to $0.01$ at $K\!=\!90\%$,
while $\bar A_{\mathrm{rgb}}$ rises from $0.50$ to $0.69$. When only
the camera is corrupted, a mirrored but smaller redistribution
appears ($\bar A_{\mathrm{rgb}}\!:0.50\!\to\!0.38$,
$\bar A_{\mathrm{lid}}\!:0.25\!\to\!0.34$). In the both-corrupted
scenario, the LiDAR attention is suppressed sharply while the RGB
attention continues to grow, consistent with the wider dynamic range
of the LiDAR variance head observed in Table~\ref{tab:per_mode_v}.

Table~\ref{tab:per_mode_v} probes the variance head one corruption
mode at a time. Twelve of the thirteen trained modes produce a
variance ratio of at least $3.2\times$ the clean baseline, with the
most severe modes (camera \emph{signal\_drop} at $37.5\times$, LiDAR
\emph{signal\_drop} at $57.9\times$, LiDAR \emph{frustum\_occlusion}
at $37.7\times$) producing the largest responses. The trust shift
$\Delta t$ is more decisive on the LiDAR side than on the camera
side ($-0.47$ vs $-0.11$ for signal\_drop), which reflects the larger
dynamic range of the LiDAR variance head. The single outlier is \emph{color\_shift}, which produces a near-flat
$1.1\times$ ratio. This weak response suggests that the current feature
extractor is relatively insensitive to this transformation. A dedicated
per-mode closed-loop analysis is needed to determine whether the weak
reliability response corresponds to negligible planning impact.

\subsection{Qualitative Analysis}\label{subsec:qualitative}

\begin{figure}[!t]
\centering
\includegraphics[width=\columnwidth]{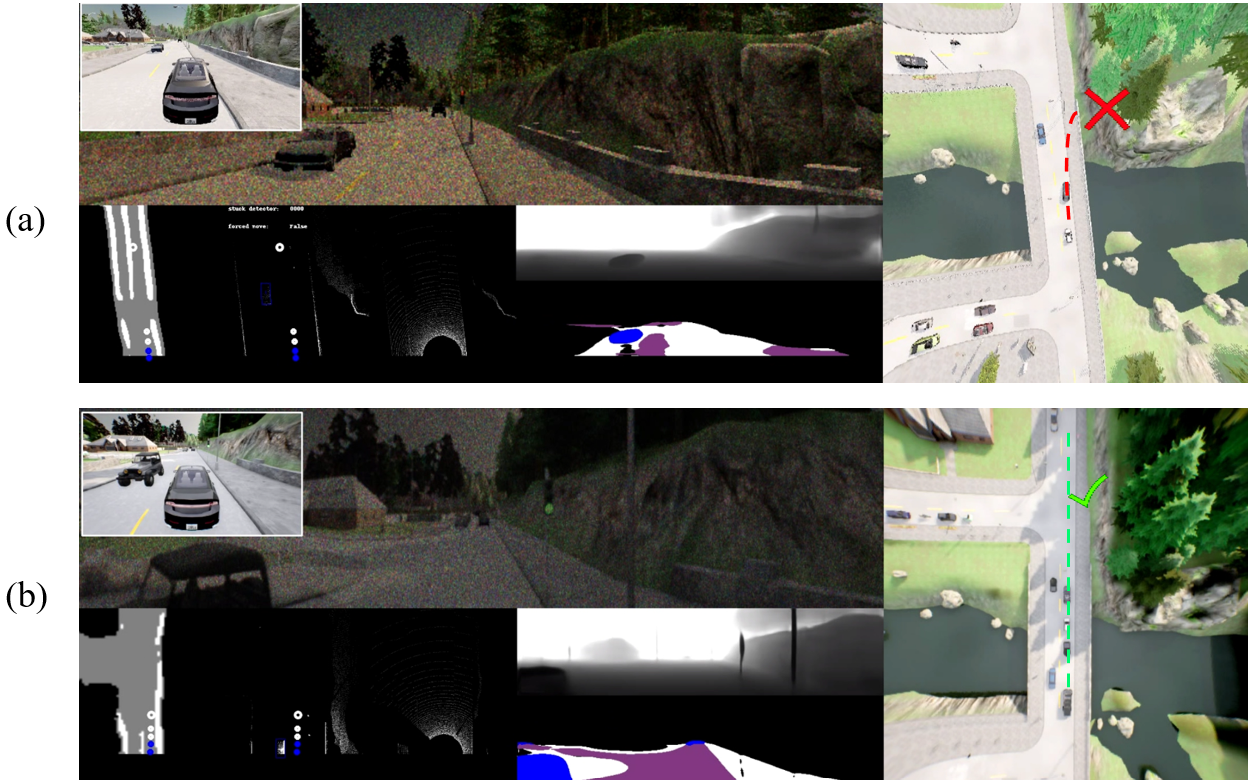}
\caption{Closed-loop comparison under the \textit{night\_lowlight}
corruption mode. (a) TransFuser baseline: the RGB semantic
segmentation labels the gravel pavement and shoulder as drivable
road; the ego exits the carriageway at the curve and the route fails
(red cross). (b) VG-SAF: the RGB segmentation confines the road
class to the carriageway and the route completes successfully (green
check).}
\label{fig:carla_qualitative}
\end{figure}

Fig.~\ref{fig:qualitative_rgb} and Fig.~\ref{fig:qualitative_lidar} illustrate 
single-frame visualization of variance-guided fusion under
asymmetric corruption. Each row shows the augmented inputs, per-task predictions, dense
reliability maps, the per-cell attention $A^{\mathrm{loc}}_m$, the
trust-scaled product $A_m$, and the predicted waypoints.
Fig.~\ref{fig:qualitative_rgb} shows a camera-side
\emph{local\_exposure\_pulse} at moderate severity. The over-exposed
band washes out foreground vehicles, degrades semantic masks and
introduces a spurious mid-range depth surface. The camera variances
rise to $4.77$ and $0.90$, while the LiDAR variances remain close to
the clean floor, so the RGB local gate suppresses the damaged region
and the fused waypoint sequence still follows the expert trajectory.
Fig.~\ref{fig:qualitative_lidar} shows the complementary LiDAR-side
\emph{frustum\_occlusion}. The missing wedge removes detections and
road surface evidence inside the masked footprint, causing the LiDAR
variance to peak at $9.30$ while the camera variances stay below
$2.09$. The LiDAR attention is attenuated inside the wedge and the
camera-supported trajectory remains aligned with the expert. Together,
Figs.~\ref{fig:qualitative_rgb} and~\ref{fig:qualitative_lidar}
illustrate that the same variance-guided mechanism works for both
camera and LiDAR faults rather than relying on a modality-specific
failure rule.

Fig.~\ref{fig:carla_qualitative} shows a closed-loop replay of one
route under \emph{night\_lowlight} corruption. Both agents share the
same backbones and the same camera input. The TransFuser baseline
veers off the drivable surface at the curve and the route fails. VG-SAF
stays on the carriageway and arrives successfully. The visualizations suggest that one contributing factor is the RGB
semantic segmentation: in the baseline the predicted
road class bleeds onto the adjacent gravel pavement, so the planner
is supplied with an inflated drivable region. In VG-SAF the
camera-side variance head elevates under the global luminance loss
and the local spatial gate attenuates the impaired cells before they
reach the fused representation. The LiDAR-derived spatial cue
therefore reaches the planner intact. This example illustrates behavior also observed qualitatively under
several corruption modes and shows how reliability gating can change the
closed-loop failure mode.
Global biases such as \emph{night\_lowlight}, \emph{motion\_blur} and
LiDAR \emph{feature\_noise} push the baseline off-road. Local masks
such as RGB \emph{local\_occlusion} and LiDAR
\emph{frustum\_occlusion} push it into collisions. VG-SAF reduces the occurrence of these failure modes in the evaluated
routes by gating impaired cells before they enter the fused
representation that drives the waypoint planner.

A presentation video summarizing the complete framework and simulation demonstrations are available at \href{https://youtu.be/ACBnq_cEt3s}{Demo Video}.

\section{Limitations and Future Directions}\label{sec:limitations}

\begin{enumerate}
\item \emph{Reliability-scale calibration and encoder blind spots.}
Table~\ref{tab:per_mode_v} and Fig.~\ref{fig:attn_vs_K} show that the
LiDAR reliability head has a larger dynamic range than the camera head
under matched corruption, partly because the camera heads use
cross-entropy supervision whereas the LiDAR detector includes a
focal-loss-based classification branch. Although the modality-specific
trust gains partially compensate for scale differences, residual
miscalibration remains because the reliability distributions differ in
shape and dynamic range. Post-hoc calibration, distribution-level
regularization, or head-specific temperature calibration could improve
comparability. Encoder-invariant faults such as \emph{color\_shift} may
also produce a weak response. A lightweight input-space shift detector
could complement this blind spot while preserving the reliability-gating
principle.
\item \emph{Saturation under total signal loss.} When $K\!=\!100\%$,
the perception encoder is pushed outside its variance-informative range.
The systemic waypoint uncertainty scale $V_{\mathrm{wp}}$ still reflects the
planning error, but the per-cell gate can no longer localize reliable
evidence. Combining VG-SAF with an external sensor-health signal or a
fallback policy would improve behavior in complete outage regimes.
\item \emph{Simulation-to-real transfer.} All quantitative results are
obtained in CARLA. Although the augmentor is physically grounded, real
sensor faults may differ in texture, temporal persistence and spatial
statistics. Robo3D and RoboBEV~\cite{cite:robo3d, cite:robobev} provide
a controlled bridge, and adapting cross-branch distillation to their
real-world corruption masks is a natural next step toward deployment on
physical vehicular platforms.
\end{enumerate}

\section{Conclusion}\label{sec:conclusion}

We have presented VG-SAF, a multimodal fusion stage for end-to-end
driving under asymmetric and spatially localized sensor degradation. The
central problem addressed in this work is that a corrupted modality is
rarely uniformly useless: damaged cells should be suppressed, but
unaffected cells of the same stream should still support planning. VG-SAF
addresses this problem through three linked mechanisms. The physically
grounded augmentor provides dense mask supervision without manual
uncertainty labels; cross-branch dense distillation converts the mask
into a monotone severity-to-reliability response; and hybrid
variance-guided attention transforms this calibrated reliability into both
local spatial suppression and cross-modal trust arbitration. A systemic waypoint uncertainty scale $V_{\mathrm{wp}}$ trained under the Laplace NLL of
Eq.~\eqref{eq:wpnll} further exposes severe or combined degradation as a safety
signal. On the Longest6 benchmark, VG-SAF exceeds TransFuser by
$10.5$, $8.5$ and $13.7$ driving-score points under camera, LiDAR and
both-modality corruption respectively, while also providing interpretable
attention and trust maps. These results show that robustness can be
improved not only by adding more sensors or larger fusion modules, but
by making the fusion operation explicitly reliability-aware at the
spatial cell level. Future extensions should combine this spatial
reliability with temporal reliability modeling and real-vehicle fault
validation, which are necessary steps toward robust continuous multimodal autonomy.

\clearpage
\setcounter{section}{0}
\setcounter{figure}{0}
\setcounter{table}{0}
\setcounter{equation}{0}
\setcounter{algorithm}{0}
\renewcommand{\thesection}{S\arabic{section}}
\renewcommand{\thefigure}{S\arabic{figure}}
\renewcommand{\thetable}{S\arabic{table}}
\renewcommand{\theequation}{S\arabic{equation}}
\renewcommand{\thealgorithm}{S\arabic{algorithm}}
\renewcommand{\theHsection}{S\arabic{section}}
\renewcommand{\theHfigure}{S\arabic{figure}}
\renewcommand{\theHtable}{S\arabic{table}}
\renewcommand{\theHequation}{S\arabic{equation}}
\providecommand{\theHalgorithm}{\thealgorithm}
\renewcommand{\theHalgorithm}{S\arabic{algorithm}}

\twocolumn[
\begin{center}
{\LARGE\bfseries Supplementary Material}\\
\vspace{0.4em}
{\large\bfseries Variance-Guided Spatial Attention Fusion for Robust
End-to-End Driving under Asymmetric Sensor Degradation}
\end{center}
\vspace{0.6em}
]

\section{Notation}\label{sm:notation}

This document collects all derivations, hyper-parameters and
implementation details that complement the main paper. We retain the
notation introduced in the main text. The corruption mask
$\mathbf{M}\!\in\![0,1]^{H\times W}$ is the augmentor's per-pixel
record of damage. The per-pixel residual and variance of a task head
are $\ell(\hat y, y)$ and $V$. The per-modality pooled variance is
$\bar V_m$, with $m\!\in\!\{\mathrm{rgb},\mathrm{lid}\}$. The local
spatial gate is $A^{\mathrm{loc}}_m$, the cross-modal trust scalar
is $t_m$, and their product is $A_m$. The systemic waypoint uncertainty is denoted by
$V_{\mathrm{wp}}$ for consistency with the implementation and figure
labels; statistically, it is a positive trajectory-level Laplace scale,
not a variance. The unified sample-level corruption severity is
$K=100\,\mathbb{E}[\mathbf{M}]\%$. The four unconstrained learnable
gains are $\gamma^{\mathrm{loc}}_m$ and $\gamma^{\mathrm{tru}}_m$; their
positive softplus-reparameterized counterparts are
$\eta^{\mathrm{loc}}_m$ and $\eta^{\mathrm{tru}}_m$.

\section{Network Architecture Details}\label{sm:arch}

\subsection{Inputs and Targets}
The simulator provides raw $480\!\times\!800$ front-camera frames,
which are resized to form the network input
$\mathbf{x}_{\mathrm{rgb}}\!\in\!\mathbb{R}^{3\times 160\times 704}$ and
then mean--variance normalized using the ImageNet statistics. The wide
aspect ratio preserves the horizontal field of view that contains lane
geometry and traffic-light cues. The LiDAR input is a voxelized
bird's-eye-view tensor
$\mathbf{x}_{\mathrm{lid}}\!\in\!\mathbb{R}^{2\times 256\times 256}$.
The first channel stores the maximum height of any LiDAR point in the
cell. The second stores the mean laser-return intensity. The grid
covers $32\!\times\!32~\mathrm{m}^2$ in front of the ego vehicle at
$8$ pixels per meter, with the sensor at the bottom-center and the
forward axis pointing upward. The third input is the navigation goal
$\mathbf{p}_{t}\!\in\!\mathbb{R}^{2}$ in the ego-vehicle frame. The
training targets are $T\!=\!4$ future ground-truth waypoints
$\mathbf{y}_{1:T}\!\in\!\mathbb{R}^{T\times 2}$, dense semantic and
depth maps for the camera, a BEV segmentation map for the LiDAR, and
per-frame bounding-box annotations.

\subsection{Encoders}
Both modality experts adopt the RegNetY backbone family
\cite{cite:regnety}. The camera backbone is initialized from
ImageNet weights and produces a feature map
$\mathbf{f}_{\mathrm{rgb}}\!\in\!\mathbb{R}^{C\times h_I\times w_I}$
with $C\!=\!512$ and $(h_I,w_I)\!=\!(5,22)$. The LiDAR backbone shares
the same architecture but is trained from scratch with a fresh
$3\!\times\!3$ stem convolution that accepts the two-channel BEV
input. The deepest stage is projected to $C\!=\!512$, yielding
$\mathbf{f}_{\mathrm{lid}}\!\in\!\mathbb{R}^{C\times h_B\times w_B}$
with $h_B\!=\!w_B\!=\!8$. A three-stage feature pyramid produces
higher-resolution heads at $16\!\times\!16$, $32\!\times\!32$ and
$64\!\times\!64$ with $64$ channels.

\subsection{Task Heads and Softplus Parameterization}
Each task head shares a decoder body with its reliability head, so the
two outputs are spatially aligned. The camera segmentation, LiDAR BEV
segmentation, and detection scales use
\begin{equation}
V = \mathrm{sp}(\rho) + \varepsilon ,\qquad
\varepsilon = 10^{-2},
\label{sm:softplus}
\end{equation}
with $\mathrm{sp}(\cdot)$ the softplus and $\rho$ the unconstrained
head output. The depth head instead predicts a standard deviation
$\sigma=0.05+\mathrm{sp}(\rho)$ and uses
$V^{\mathrm{dep}}=\sigma^2$. Softplus is smooth everywhere and
asymptotically linear for large positive inputs.

The camera expert produces a seven-class segmentation map with
reliability scale $V^{\mathrm{seg}}$ and a normalized depth map with
variance $V^{\mathrm{dep}}$. The LiDAR expert produces a three-class
BEV segmentation map with scale $V^{\mathrm{bev}}$ and a CenterNet
detection head~\cite{cite:centernet} with a per-pixel reliability scale
$V^{\mathrm{det}}$.

\subsection{Fusion Projection and Waypoint Head}
The two gated feature maps are pooled by global average and
concatenated, then projected back to width $C\!=\!512$ by a fusion
projection:
\begin{equation}
\mathbf{z}_{\mathrm{fuse}} = \mathrm{FFN}_{\mathrm{fuse}}\!\Bigl(
\bigl[\mathrm{GAP}(\tilde{\mathbf{f}}_{\mathrm{rgb}});\,
\mathrm{GAP}(\tilde{\mathbf{f}}_{\mathrm{lid}})\bigr]\Bigr) .
\label{sm:zfuse}
\end{equation}
The waypoint head is the autoregressive GRU
of~\cite{cite:gru, cite:transfuser}. The recurrence is
\begin{align}
\mathbf{h}_{0} &= \mathrm{MLP}_{\mathrm{wp}}(\mathbf{z}_{\mathrm{fuse}}) ,\\
\mathbf{h}_{t} &= \mathrm{GRU}\!\bigl([\hat{\mathbf{y}}_{t-1};\mathbf{p}_{t}],\,\mathbf{h}_{t-1}\bigr) ,\\
\hat{\mathbf{y}}_{t} &= \hat{\mathbf{y}}_{t-1} + \mathbf{W}_{o}\mathbf{h}_{t},
\label{sm:wpstep}
\end{align}
with $\hat{\mathbf{y}}_{0}\!=\!\mathbf{0}$ and a learnable read-out
$\mathbf{W}_{o}\!\in\!\mathbb{R}^{2\times D}$. The systemic
uncertainty head consumes $\mathbf{z}_{\mathrm{fuse}}$ together with
two scalar variance summaries
$s^{V,\mathrm{wp}}_{\mathrm{rgb}},s^{V,\mathrm{wp}}_{\mathrm{lid}}$:
\begin{equation}
s^{V,\mathrm{wp}}_{m}=\max_{u,v}\bar V_m(u,v),\qquad m\in\{\mathrm{rgb},\mathrm{lid}\},
\label{sm:sVwp}
\end{equation}
\begin{equation}
V_{\mathrm{wp}} = \mathrm{sp}\!\bigl(\mathrm{MLP}([\mathbf{z}_{\mathrm{fuse}};
s^{V,\mathrm{wp}}_{\mathrm{rgb}}; s^{V,\mathrm{wp}}_{\mathrm{lid}}])\bigr)
+ \varepsilon .
\label{sm:vwp}
\end{equation}
The worst-case max-pool is the natural aggregator for the safety
alarm. A single region of high uncertainty, such as a fully
occluded LiDAR frustum, immediately inflates $V_{\mathrm{wp}}$.
The planning objective matches the Laplace head loss of
Eq.~\eqref{sm:lapnll} and Eq.~(12) of the main paper:
\begin{equation}
\mathcal{L}_{\mathrm{wp}} = \frac{1}{T}\sum_{t=1}^{T}
\left(
\frac{\|\hat{\mathbf{y}}_t-\mathbf{y}_t\|_1}{V_{\mathrm{wp}}}
+ 2\log V_{\mathrm{wp}}
\right),
\label{sm:wpnll}
\end{equation}
where $V_{\mathrm{wp}}$ is a positive Laplace scale shared across all
$T$ steps and both waypoint coordinates. The factor
$2\log V_{\mathrm{wp}}$ accounts for the two independent coordinates;
the coordinate residuals are aggregated by the $\ell_1$ norm.

\section{Augmentation Catalog}\label{sm:catalog}

\begin{figure*}[!t]
\centering
\includegraphics[width=\textwidth]{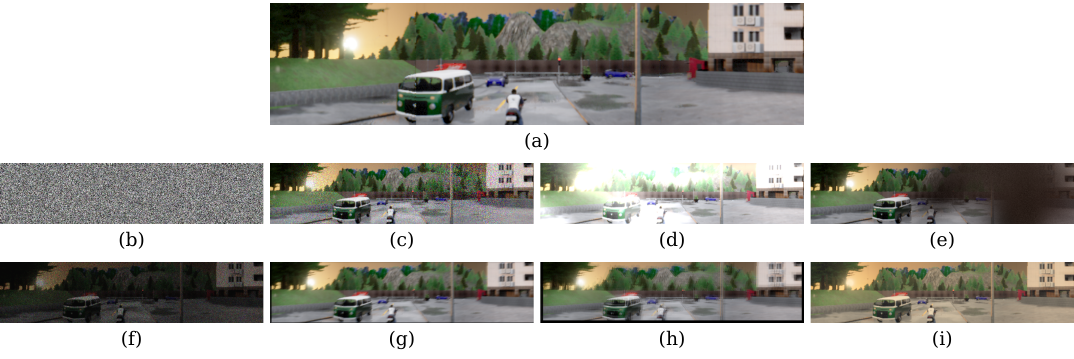}
\caption{Camera fault catalog. Panel (a) shows an unperturbed
front-camera frame from the Longest6 validation split. Panels
(b)--(i) display the eight trained corruption modes applied
independently to (a). (b) \emph{signal\_drop}; (c) \emph{local\_noise};
(d) \emph{local\_exposure\_pulse}; (e) \emph{local\_occlusion};
(f) \emph{night\_lowlight}; (g) \emph{motion\_blur}; (h) \emph{ghosting};
(i) \emph{color\_shift}. Modes (b), (f), (g), (h), (i) carry
\emph{global} masks; (c), (d), (e) carry \emph{local} masks.}
\label{fig:rgb_catalog}
\end{figure*}

\begin{figure*}[!t]
\centering
\includegraphics[width=0.8\textwidth]{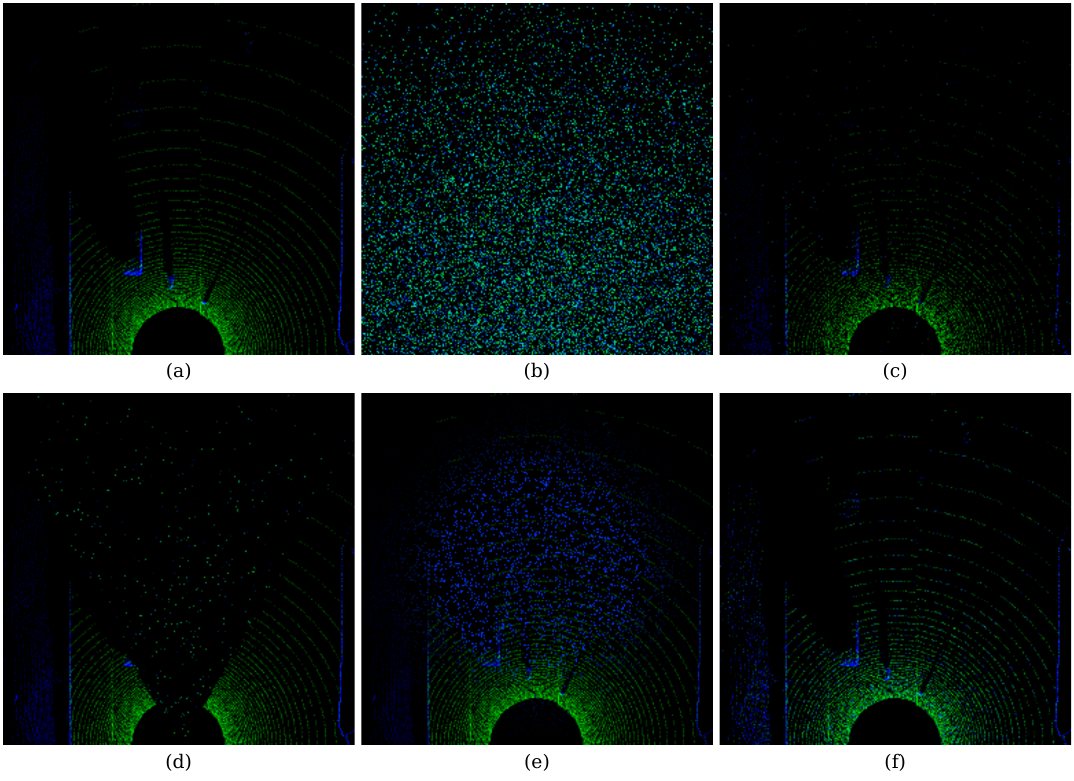}
\caption{LiDAR fault catalog. Panel (a) shows the unperturbed BEV
histogram for the same frame as Fig.~\ref{fig:rgb_catalog}(a).
Panels (b)--(f) display the five trained LiDAR corruption modes.
(b) \emph{signal\_drop}; (c) \emph{range\_dropout};
(d) \emph{frustum\_occlusion}; (e) \emph{local\_speckle};
(f) \emph{feature\_noise}. Modes (b), (c), (f) carry \emph{global}
masks; (d) and (e) carry \emph{local} masks.}
\label{fig:lidar_catalog}
\end{figure*}

This section gives the expanded corruption catalog summarized in
Section~III-C of the main paper. Tables~\ref{tab:rgb_modes}
and~\ref{tab:lidar_modes} record the real-world analogue,
description, spatial scope (global or local), and training range of
the unified severity $K$ for every mode. Figs.~\ref{fig:rgb_catalog}
and~\ref{fig:lidar_catalog} show one realization of every mode on
a single validation frame.

The same physical catalog defines the family of faults considered by
this study, but the training and evaluation realizations are separated.
Training samples corruptions online with phase-dependent probabilities,
while evaluation uses held-out Longest6 routes, independent random seeds
and route-level persistent severities that are not reused from training
mini-batches. The evaluation additionally performs severity extrapolation beyond
the training ranges in Tables~\ref{tab:rgb_modes} and
\ref{tab:lidar_modes}; this explains the larger $K$ values reported in
Table~II of the main paper. The generic ablation uses uniform random
masks and elliptical occlusions that do not exactly match a named mode.
These tests probe robustness to unseen severities and mask geometries,
but they do not establish general real-world out-of-distribution
detection; real-vehicle validation remains future work.

\begin{table*}[!t]
\centering
\renewcommand{\arraystretch}{1.12}
\caption{Camera fault modes implemented by the proposed augmentor.}
\label{tab:rgb_modes}
\footnotesize
\begin{tabularx}{\textwidth}{@{}l >{\hsize=0.5\hsize\raggedright\arraybackslash}X >{\hsize=1.5\hsize\raggedright\arraybackslash}X c c@{}}
\toprule
\textbf{ID \& Mode} & \textbf{Real-world analogue} & \textbf{Description} & \textbf{Scope} & \textbf{Training $K$} \\
\midrule
 1 \texttt{signal\_drop} & power loss / glare / snow & Frame-wide blackout, whiteout saturation, or luminance-correlated snow with chromatic jitter. & Global & $100\%$ \\ \addlinespace
 2 \texttt{local\_noise} & low-light ISO patches & 2--6 Gaussian noise patches blending Poisson photon noise and additive Gaussian read-out noise. & Local  & $10$--$40\%$ \\ \addlinespace
 3 \texttt{local\_exp}* & specular flare / bloom & 1--3 localized over-exposure spots with directional bloom and gamma--gain saturation. & Local  & $10$--$45\%$ \\ \addlinespace
 4 \texttt{local\_occ}* & lens debris / mud splash & 2--4 alpha-blended elliptical occluders with edge darkening, mimicking foreign objects. & Local  & $5$--$30\%$ \\ \addlinespace
 5 \texttt{night\_low}* & dusk / night driving & Frame-wide gain reduction, non-linear gamma, warm shift, and contrast roll-off. & Global & $30$--$65\%$ \\ \addlinespace
 6 \texttt{motion\_blur} & ego-motion / vibration & Directional convolution with a randomly oriented kernel, simulating ego-motion or shake. & Global & $23$--$68\%$ \\ \addlinespace
 7 \texttt{ghosting} & lens reflection / image persistence & 2--3 shifted, translucent ghost copies; pixels outside the original frame are marked invalid. & Global & $15$--$35\%$ \\ \addlinespace
 8 \texttt{color\_shift} & WB miscalibration & Affine color transformation with warm/cool shift, contrast roll-off, and gamma curves. & Global & $1$--$7\%$ \\
\bottomrule
\multicolumn{5}{l}{* \textit{Shortened names: local\_exposure\_pulse (3), local\_occlusion (4), night\_lowlight (5).}}
\end{tabularx}
\end{table*}

\begin{table*}[!t]
\centering
\renewcommand{\arraystretch}{1.12}
\caption{LiDAR fault modes implemented by the proposed augmentor.}
\label{tab:lidar_modes}
\footnotesize
\begin{tabularx}{\textwidth}{@{}l >{\hsize=0.5\hsize\raggedright\arraybackslash}X >{\hsize=1.5\hsize\raggedright\arraybackslash}X c c@{}}
\toprule
\textbf{ID \& Mode} & \textbf{Real-world analogue} & \textbf{Description} & \textbf{Scope} & \textbf{Training $K$} \\
\midrule
1 \texttt{signal\_drop}   & jamming / spurious returns & Frame-wide jamming represented by spurious returns following range-conditioned sparsity. & Global & $100\%$ \\ \addlinespace
2 \texttt{range\_drop}*   & rain / fog / absorption & Distance-dependent point dropping with near-sensor false returns simulating dome-side backscatter. & Global & $20$--$55\%$ \\ \addlinespace
3 \texttt{frustum\_occ}*  & mud / debris on dome & Angular wedge of complete blockage, populated with sparse near-sensor false returns. & Local  & $8$--$25\%$ \\ \addlinespace
4 \texttt{local\_speckle} & airborne dust / insects & Sum of 2--5 oriented Gaussian blobs gated to empty cells, modulated by sparse Bernoulli patterns. & Local  & $5$--$20\%$ \\ \addlinespace
5 \texttt{feature\_noise} & measurement / channel noise & Channel-wise Gaussian noise added to active voxels, modeling measurement or feature-channel disturbance. & Global & $20$--$60\%$ \\
\bottomrule
\multicolumn{5}{l}{* \textit{Shortened names: range\_dropout (2), frustum\_occlusion (3).}}
\end{tabularx}
\end{table*}

The Bernoulli noise injectors in the LiDAR augmentor are modulated by
$1\!-\!r$, where $r\!\in\![0,1]$ is the normalized distance from the
sensor origin. Peak densities are capped between $1\%$ and $3\%$.
This restores the approximate inverse-range falloff that real LiDAR
ray density obeys and prevents near-field speckles from dominating the
severity mask.

\section{Detailed Training Losses}\label{sm:losses}

\subsection{Heteroscedastic Loss Attenuation}
The regression and classification heads share the loss-attenuation form
\begin{equation}
\mathcal{L}_{\mathrm{het}}(\hat y,y;V)=a_k\!\left(
\frac{\ell_k(\hat y,y)}{V}+\log V\right),
\label{sm:gnll}
\end{equation}
where $a_k=\tfrac12$ and $\ell_k=(\hat y-y)^2$ give the Gaussian depth
NLL. For segmentation and detection heatmaps, $\ell_k$ is cross-entropy
or focal loss and Eq.~\eqref{sm:gnll} is used as an uncertainty-weighted
classification surrogate. For this generic attenuated form, the
stationary scale is proportional to the local supervised loss. The
sparse regression branches of the CenterNet detection head use the
Laplace counterpart
\begin{equation}
\mathcal{L}_{\mathrm{Lap}}(\hat{y},y;V) = |\hat{y}-y|/V + \log V ,
\label{sm:lapnll}
\end{equation}
which arises from the same maximum-likelihood derivation under a
Laplace observation model and matches the $L_1$ magnitude already
used by CenterNet.

\subsection{Dual-Branch Mask-Weighted Supervision}
For Phase 2 onward, the encoder ingests a concatenated batch
$[\mathrm{clean}\,|\,\mathrm{noisy}]$. We denote by
$(\hat y_c, V_c)$ and $(\hat y_n, V_n)$ the clean and noisy halves of
each prediction. Let $\mathbf{W}\!=\!\mathbf{1}-\mathbf{M}$ be the
complementary \emph{trustworthiness} mask. The per-task supervised
loss is
\begin{equation}
\mathcal{L}_{\mathrm{sup}}
  = \mathcal{L}_{\mathrm{het}}(\hat y_c,y;V_c)
  + \lambda_{\mathrm{sup}}\,\mathbf{W}\!\odot\!\mathcal{L}_{\mathrm{het}}(\hat y_n,y;V_n) ,
\label{sm:sup}
\end{equation}
with $\lambda_{\mathrm{sup}}\!=\!1$ in the reference configuration.
The mask $\mathbf{W}$ retains the full task gradient on clean cells and
progressively reduces it as the noisy observation loses reliable
evidence. The ground-truth label itself remains valid. For the CenterNet branches, the mean
reduction is replaced by the standard $\mathrm{sum}(\cdot)/\mathrm{af}$
normalization, where $\mathrm{af}$ counts positive object-center
cells.

\subsection{Confidence-Aware Consistency}
A mask-weighted consistency term ties the noisy branch back to the
clean branch on uncorrupted pixels:
\begin{align}
\mathcal{L}^{\mathrm{cls}}_{\mathrm{con}}
&= \mathbf{W}\!\odot\!\mathrm{KL}\!\bigl(\mathbf{P}_n \,\|\, \mathrm{sg}(\mathbf{P}_c)\bigr) ,\label{sm:concls}\\
\mathcal{L}^{\mathrm{reg}}_{\mathrm{con}}
&= \mathbf{W}\!\odot\!\bigl(\hat y_n - \mathrm{sg}(\hat y_c)\bigr)^{2},\label{sm:conreg}
\end{align}
where $\mathrm{sg}(\cdot)$ is the stop-gradient operator and
$\mathbf{P}\!=\!\mathrm{softmax}(\mathbf{z})$ is the per-pixel class
probability vector. The mask is essential: without it, the
consistency term would penalise corrupted pixels for differing from
their clean counterparts, which is the opposite of what the reliability head requires.

\subsection{Cross-Branch Dense Distillation}
The novel calibration term sets a per-pixel target on the
log-reliability scale of the noisy branch that is linear in the local
corruption severity:
\begin{align}
\log V_{\mathrm{tgt}} &= \mathrm{sg}(\log V_c) + \alpha\mathbf{M},\nonumber\\
\mathcal{L}_{\mathrm{cal}} &= \mathrm{SmoothL1}(\log V_n, \log V_{\mathrm{tgt}}).
\label{sm:cal}
\end{align}
The scaling constant $\alpha$ is decoupled per head. Both camera
heads and the LiDAR BEV-segmentation head sit on a cross-entropy
baseline of order $10^{-1}$, so $\alpha_{\mathrm{seg}}\!=\!3$ produces
a $20\!\times$ ratio at $\mathbf{M}\!=\!1$. The focal-loss-squashed
CenterNet baseline of order $10^{-2}$ requires $\alpha_{\det}\!=\!4.6$
for the matching $100\!\times$ ratio.

\subsection{Total Perception Loss}
The per-modality training objective at epoch $e$ is
\begin{equation}
\begin{aligned}
\mathcal{L}_{\mathrm{percep}}(e) &= \sum_{k\in\mathcal{T}}
  \mathcal{L}^{(k)}_{\mathrm{sup}}\\
&\quad+ \lambda_{\mathrm{con}}(e)\sum_{k}\mathcal{L}^{(k)}_{\mathrm{con}}
+ \lambda_{\mathrm{cal}}(e)\sum_{k}\mathcal{L}^{(k)}_{\mathrm{cal}},
\end{aligned}
\label{sm:totalpercep}
\end{equation}
where $\mathcal{T}\!=\!\{\mathrm{seg},\mathrm{dep}\}$ for the camera
expert and $\mathcal{T}\!=\!\{\mathrm{bev},\mathrm{det}\}$ for the
LiDAR expert. The term $\mathcal{L}^{(k)}_{\mathrm{sup}}$ is the
mask-weighted clean--noisy supervision of Eq.~\eqref{sm:sup}, so
the noisy-branch task supervision is progressively reduced as corruption severity increases.

\section{Fusion-Stage Details}\label{sm:fusion}

\subsection{Variance Normalizers and Pooling}
The four per-task variance maps carry the same qualitative notion of
cell-level reliability signal but live on different magnitude scales.
The two cross-entropy heads sit on a baseline of order $10^{-1}$. The
Gaussian depth head sits on a squared-error baseline of order
$10^{-2}$. The focal-loss detection head is squashed by construction
to a baseline of order $10^{-2}$. We normalize each summand by a
clean-data empirical mean
$n^{\mathrm{seg,rgb}},n^{\mathrm{dep}},n^{\mathrm{seg,lid}},n^{\mathrm{det}}$
so that each task-specific variance channel has a unit-mean baseline
before summation:
\begin{equation}
V_{\mathrm{rgb}} = \frac{V^{\mathrm{seg}}}{n^{\mathrm{seg,rgb}}}
 + \frac{V^{\mathrm{dep}}}{n^{\mathrm{dep}}},\quad
V_{\mathrm{lid}} = \frac{V^{\mathrm{bev}}}{n^{\mathrm{seg,lid}}}
 + \frac{V^{\mathrm{det}}}{n^{\mathrm{det}}}.
\label{sm:vsum}
\end{equation}
The normalizers are scalars, are computed once on a clean validation
split, and are frozen throughout fusion training. Without
normalization, the cross-entropy summands would dominate by an order
of magnitude on every clean frame and force the attention gains to
absorb a static per-modality offset.

\subsection{Softplus Reparameterization of the Gains}
The four learnable scalars
$\gamma^{\mathrm{loc}}_{\mathrm{rgb}},\gamma^{\mathrm{loc}}_{\mathrm{lid}},
 \gamma^{\mathrm{tru}}_{\mathrm{rgb}},\gamma^{\mathrm{tru}}_{\mathrm{lid}}$
are unconstrained. To prevent any of them from becoming negative during transient optimizer steps, we softplus-reparameterize
each at the start of every forward pass:
\begin{equation}
\eta^{r}_{m} = \mathrm{sp}(\gamma^{r}_{m}) + 10^{-4},
\quad m\!\in\!\{\mathrm{rgb},\mathrm{lid}\},\;
r\!\in\!\{\mathrm{loc},\mathrm{tru}\}.
\label{sm:gammasp}
\end{equation}
The floor $10^{-4}$ guarantees $\eta^{r}_{m}\!>\!0$ while still
allowing linear growth for large positive $\gamma^{r}_{m}$.

\subsection{Trust Floor}
The cross-modal trust softmax of the main paper is followed by an
affine map that bounds the output:
\begin{equation}
\tilde t_m = \tau + (1-2\tau) t_m ,\qquad \tau\!\in\![0,\tfrac12) ,
\label{sm:trustfloor}
\end{equation}
so that $\tilde t_m\!\in\![\tau,1-\tau]$ for both modalities. We use
$\tau\!=\!0.15$, which corresponds to a worst-case modality-trust
ratio of $5.7{:}1$. Setting $\tau\!=\!0$ recovers the unclamped
softmax. The trust floor prevents the over-suppression of an
otherwise spatially-recoverable modality, matching the safety rationale
summarized in the main paper.

\subsection{Trust-Balance Regularizer}
The cross-modal trust is intentionally underdetermined on no-fault
frames. To prevent silent drift toward a single dominant modality,
we add a lightweight regularizer that pulls the trust split back
toward the prior on samples that received no augmentation:
\begin{equation}
\mathcal{L}_{\mathrm{tb}}
 = \frac{1}{|\mathcal{I}_c|}\sum_{i\in\mathcal{I}_c}
 \bigl(t^{(i)}_{\mathrm{rgb}} - \tfrac12\bigr)^{2},
\label{sm:tbreg}
\end{equation}
where $\mathcal{I}_c$ is the set of mini-batch samples with
$\beta^{(i)}_{\mathrm{rgb}}\!=\!\beta^{(i)}_{\mathrm{lid}}\!=\!0$.
The regularizer is zero by construction whenever no clean samples
appear in the mini-batch. The complete fusion-stage objective is
\begin{equation}
\mathcal{L}_{\mathrm{fusion}}
 = \mathcal{L}_{\mathrm{wp}} + \lambda_{\mathrm{tb}}\,\mathcal{L}_{\mathrm{tb}},
\label{sm:fusionloss}
\end{equation}
with $\lambda_{\mathrm{tb}}\!=\!0.5$ in the reference configuration.

\subsection{Quantile-versus-Maxpool Aggregation}
The trust softmax consumes the $25$th percentile $Q_{0.25}(\bar V_m)$
of the variance map. The systemic head consumes the spatial maximum
$\max(\bar V_m)$. The two aggregators serve different goals. The
trust softmax is an arbitration signal that must remain insensitive
to isolated noisy cells, since these are already attenuated by the
local gate. A lower-quartile aggregator asks whether the modality
still has enough reliable cells to be worth trusting at all. The
systemic head, in contrast, is a safety alarm that should fire as
soon as any sufficiently large region of either modality is severely
corrupted. The maximum is the natural aggregator for that role because it preserves
the worst-case local evidence.

\section{Training Curricula}\label{sm:curriculum}

\subsection{Perception-Stage Curriculum}
The reference configuration uses $P_{1}\!=\!45$, $P_{2}\!=\!65$,
$P_{3}\!=\!90$, $N_{\mathrm{ep}}\!=\!140$,
$\lambda^{\star}_{\mathrm{con}}\!=\!0.5$,
$\lambda^{\star}_{\mathrm{cal}}\!=\!1.0$, and
$\Delta P_2\!=\!P_2\!-\!P_1\!=\!20$. The four phases proceed as
follows.

\textit{Phase 1 (warm-up).} For epochs $1\!\le\!e\!\le\!P_1$, the
encoder receives single-branch inputs with
$\mathrm{p}_{\mathrm{aug}}\!=\!0.20$. The strongest mode is excluded
for the camera and capped at a $5\%$ frame share for the LiDAR. All
calibration weights are zero, so the loss reduces to the supervised heteroscedastic loss.

\textit{Phase 2 (modality balancing and ramp-up).} For
$P_1\!<\!e\!\le\!P_2$, each batch is split into clean and noisy
halves under a uniform mode mix. The consistency and calibration weights
ramp from zero on \emph{staggered} schedules:
\begin{align}
\lambda_{\mathrm{con}}(e) &=
  \lambda^{\star}_{\mathrm{con}}\!\cdot\!
  \min\!\Bigl(\tfrac{e-P_1}{f_{\mathrm{con}}\Delta P_2},1\Bigr) ,\label{sm:rampcon}\\
\lambda_{\mathrm{cal}}(e) &=
  \lambda^{\star}_{\mathrm{cal}}\!\cdot\!
  \mathrm{clip}\!\Bigl(\tfrac{e-P_1-f_{\mathrm{cal}}\Delta P_2}{(1-f_{\mathrm{cal}})\Delta P_2},0,1\Bigr) ,\label{sm:rampcal}
\end{align}
where $\mathrm{clip}(x,a,b)\!=\!\min(\max(x,a),b)$. We use
$f_{\mathrm{con}}\!=\!0.5$ and $f_{\mathrm{cal}}\!=\!0.3$, so the
calibration weight begins to ramp after $30\%$ of Phase~2, while the
consistency weight reaches its maximum at the midpoint of the phase.

\textit{Phase 3 (uniform-mode consolidation).} For
$P_2\!<\!e\!\le\!P_3$, the dual-branch regime continues under the
uniform mode mix at full calibration weights. A few additional epochs
of stable exposure absorb the transition to Phase 4 smoothly.

\textit{Phase 4 (hard-fault mining).} For $e\!>\!P_3$, the mode
distribution shifts toward the hard physical failures (motion blur,
low-light operation, ghosting for the camera; frustum occlusion and
range-dependent dropout for the LiDAR). The late training budget is
spent on the corner cases for which a calibrated reliability head
produces tangible safety value.

\subsection{BatchNorm Freeze}
From the start of Phase 2 onward, every batch is a clean--noisy
mixture. The BatchNorm running statistics therefore drift away from
either pure distribution. At inference, the input is
single-distribution, and the mismatch is most damaging under severe
corruptions: the activation magnitude collapses, the variance-head
pre-activation $\rho$ drifts deep into the negative region, and
$V$ saturates at the floor $\varepsilon$ across the whole frame. The
result is a collapsed or unresponsive variance head. We resolve this by
forcing every BatchNorm layer into evaluation mode from the
beginning of Phase 2. The affine parameters remain trainable; only
the running-statistics update is disabled, preserving adaptation while
preventing the calibration collapse observed under mixed batches.

\subsection{Fusion-Stage Curriculum}
The fusion curriculum uses $P^f_1\!=\!20$, $P^f_2\!=\!40$,
$P^f_3\!=\!60$, and a total of $100$ epochs.

\textit{Phase 1.} $\beta_{\mathrm{rgb}}\!=\!\beta_{\mathrm{lid}}\!=\!0$
throughout. The GRU and the systemic uncertainty head learn the base
navigation behavior on clean inputs only.

\textit{Phase 2.} Each sample draws $(\beta_{\mathrm{rgb}},\beta_{\mathrm{lid}})$
uniformly from $\{(0,0),(1,0),(0,1)\}$. Single-modality faults
exercise the attention gains without yet stressing the systemic head.

\textit{Phase 3.} The two flags are sampled independently with
$\Pr(\beta_m\!=\!1)\!=\!0.5$. Approximately a quarter of the samples
contain double faults that stress the systemic uncertainty head $V_{\mathrm{wp}}$.

\textit{Phase 4.} The Bernoulli sampling continues, but the
augmentor mode mix switches to the hard distribution of
Section~\ref{sm:curriculum}\,A. $V_{\mathrm{wp}}$ is calibrated to
increase under severe combined faults and severities outside the training ranges.

\section{End-to-End Training and Forward-Pass Algorithms}\label{sm:algorithms}
Algorithm~\ref{alg:workflow} summarizes the two training stages.
Algorithm~\ref{alg:vgsaf_forward} is a pure forward-inference procedure,
whereas Algorithm~\ref{alg:fusion_train} adds the ground truth and losses
required only during fusion training.

\begin{algorithm}[!t]
\caption{End-to-End VG-SAF Training Workflow}
\label{alg:workflow}
\footnotesize
\begin{algorithmic}[1]
\REQUIRE Dataset $\mathcal{D}$, augmentors $\mathcal{A}_{\mathrm{rgb}},\mathcal{A}_{\mathrm{lid}}$, perception epochs $N^{p}_{\mathrm{ep}}$, fusion epochs $N^{f}_{\mathrm{ep}}$.
\STATE \textbf{Stage 1: train per-modality perception experts.}
\FOR{$m\in\{\mathrm{rgb},\mathrm{lid}\}$}
  \FOR{$e=1,\ldots,N^{p}_{\mathrm{ep}}$}
    \STATE Select the curriculum phase and $(p_{\mathrm{aug}},\lambda_{\mathrm{con}}(e),\lambda_{\mathrm{cal}}(e))$.
    \IF{$e=P_1+1$} \STATE Freeze all BatchNorm running statistics. \ENDIF
    \FORALL{minibatches}
      \STATE Form a single branch in Phase~1 or a $[\mathrm{clean}|\mathrm{noisy}]$ batch thereafter; obtain $\mathbf M$.
      \STATE Compute Eqs.~\eqref{sm:gnll}--\eqref{sm:cal} and optimize Eq.~\eqref{sm:totalpercep}.
    \ENDFOR
  \ENDFOR
\ENDFOR
\STATE \textbf{Stage 2: train fusion components on frozen experts.}
\STATE Compute the per-head scalar normalizers $n^{\mathrm{seg,rgb}},n^{\mathrm{dep}},n^{\mathrm{seg,lid}},n^{\mathrm{det}}$ on clean validation data.
\FOR{$e=1,\ldots,N^{f}_{\mathrm{ep}}$}
  \FORALL{minibatches}
    \STATE Sample fault flags, apply the corresponding augmentors, and run Algorithm~\ref{alg:fusion_train}.
  \ENDFOR
\ENDFOR
\end{algorithmic}
\end{algorithm}

\begin{algorithm}[!t]
\caption{VG-SAF Forward Inference}
\label{alg:vgsaf_forward}
\footnotesize
\begin{algorithmic}[1]
\REQUIRE Inputs $(\mathbf{x}_{\mathrm{rgb}},\mathbf{x}_{\mathrm{lid}},\mathbf{p}_{t})$, scalar normalizers $\{n^{\bullet}\}$, learned gains, and frozen perception experts.
\STATE Obtain $\mathbf f_m$ and $V^{\mathrm{seg,dep,bev,det}}$ from the frozen experts.
\STATE Compute $V_m$ by Eq.~\eqref{sm:vsum} and $\bar V_m\leftarrow\mathrm{maxpool}(V_m)$.
\STATE Compute positive gains, $A^{\mathrm{loc}}_m$, trust weights, the trust floor, and $A_m=\tilde t_mA^{\mathrm{loc}}_m$.
\STATE Form $\tilde{\mathbf f}_m=A_m\odot\mathbf f_m$ and $\mathbf z_{\mathrm{fuse}}$ by Eq.~\eqref{sm:zfuse}.
\STATE Roll out the GRU and compute $V_{\mathrm{wp}}$ by Eq.~\eqref{sm:vwp}.
\RETURN $\hat{\mathbf y}_{1:T},V_{\mathrm{wp}},(t_{\mathrm{rgb}},t_{\mathrm{lid}}),\bar V_m,A_m$.
\end{algorithmic}
\end{algorithm}

\begin{algorithm}[!t]
\caption{Fusion-Stage Training Step}
\label{alg:fusion_train}
\footnotesize
\begin{algorithmic}[1]
\REQUIRE A minibatch with targets $\mathbf y_{1:T}$ and fault flags $(\beta_{\mathrm{rgb}},\beta_{\mathrm{lid}})$.
\STATE Run Algorithm~\ref{alg:vgsaf_forward} for every minibatch sample.
\STATE Compute $\mathcal L_{\mathrm{wp}}$ by Eq.~\eqref{sm:wpnll} with $2\log V_{\mathrm{wp}}$.
\STATE Form $\mathcal I_c=\{i:\beta^{(i)}_{\mathrm{rgb}}=\beta^{(i)}_{\mathrm{lid}}=0\}$ and compute $\mathcal L_{\mathrm{tb}}$ by Eq.~\eqref{sm:tbreg}.
\STATE Back-propagate $\mathcal L_{\mathrm{fusion}}=\mathcal L_{\mathrm{wp}}+\lambda_{\mathrm{tb}}\mathcal L_{\mathrm{tb}}$.
\end{algorithmic}
\end{algorithm}

\section{Compute and Reproducibility}\label{sm:compute}

Stage~1 (perception) is trained for $140$ epochs on two NVIDIA
RTX~6000 Ada GPUs at a wall-clock time of approximately $30$ hours per
modality expert. Stage~2 (fusion) is trained for $100$ epochs on the same
hardware at a wall-clock time of approximately $18$ hours. Inference runs
at $34\,\mathrm{ms}$ per frame on a single RTX~6000 Ada at the input
resolutions reported in the main paper. The optimizer is AdamW with
a cosine-decay schedule; the perception-stage learning rate is
$10^{-3}$ and the fusion-stage learning rate is $5\!\times\!10^{-4}$.
Per-GPU batch sizes are $100$ (camera), $256$ (LiDAR), and $256$
(fusion).
Gradient clipping is set to $1.0$. The full training and evaluation
code, the augmentor implementation, the corruption-mask generator and
the trained checkpoints will be released upon acceptance to support
reproduction.

\IEEEtriggeratref{45}


\begin{thebibliography}{54}

\bibitem{cite:chen_survey}
L. Chen, P. Wu, K. Chitta, B. Jaeger, A. Geiger, and H. Li, ``End-to-end autonomous driving: Challenges and frontiers,'' \emph{IEEE Trans. Pattern Anal. Mach. Intell.}, vol. 46, no. 12, pp. 10164--10183, 2024.

\bibitem{cite:nuscenes}
H. Caesar, V. Bankiti, A. H. Lang, S. Vora, V. E. Liong, Q. Xu, A. Krishnan, Y. Pan, G. Baldan, and O. Beijbom, ``nuScenes: A multimodal dataset for autonomous driving,'' in \emph{Proc. IEEE/CVF Conf. Comput. Vis. Pattern Recognit. (CVPR)}, 2020, pp. 11618--11628.

\bibitem{cite:carla}
A. Dosovitskiy, G. Ros, F. Codevilla, A. Lopez, and V. Koltun, ``CARLA: An open urban driving simulator,'' in \emph{Proc. Conf. Robot Learn. (CoRL)}, 2017, pp. 1--16.

\bibitem{cite:leaderboard}
CARLA Team, ``CARLA autonomous driving leaderboard,'' 2022. [Online]. Available: \url{https://leaderboard.carla.org/}. Accessed: Jul. 2026.

\bibitem{cite:fusion_transformer}
A. Prakash, K. Chitta, and A. Geiger, ``Multi-modal fusion transformer for end-to-end autonomous driving,'' in \emph{Proc. IEEE/CVF Conf. Comput. Vis. Pattern Recognit. (CVPR)}, 2021, pp. 7077--7087.

\bibitem{cite:tvt_vision_hrl}
J. Wang, H. Sun, and C. Zhu, ``Vision-based autonomous driving: A hierarchical reinforcement learning approach,'' \emph{IEEE Trans. Veh. Technol.}, vol. 72, no. 9, pp. 11213--11226, Sep. 2023.

\bibitem{cite:tvt_netroller}
R. Xin, H. Liu, X. Mei, W. Liu, M. Ye, Z. Chen, and J. Ma, ``NetRoller: Interfacing general and specialized models for end-to-end autonomous driving,'' \emph{IEEE Trans. Veh. Technol.}, early access, 2025.

\bibitem{cite:tvt_openmpd}
X. Zhang, Z. Li, Y. Gong, D. Jin, J. Li, L. Wang, Y. Zhu, and H. Liu, ``OpenMPD: An open multimodal perception dataset for autonomous driving,'' \emph{IEEE Trans. Veh. Technol.}, vol. 71, no. 3, pp. 2437--2447, Mar. 2022.

\bibitem{cite:tvt_radar_lidar}
L. Wang, X. Zhang, J. Li, B. Xv, R. Fu, H. Chen, L. Yang, D. Jin, and L. Zhao, ``Multi-modal and multi-scale fusion 3D object detection of 4D radar and LiDAR for autonomous driving,'' \emph{IEEE Trans. Veh. Technol.}, vol. 72, no. 5, pp. 5628--5641, May 2023.

\bibitem{cite:tvt_multiruler}
Y. Xu, B. Li, Z. Zhu, W. Liu, G. Jia, G. Han, and X. Li, ``MultiRuler: A multi-dimensional resource modeling method for embedded intelligent systems of autonomous driving,'' \emph{IEEE Trans. Veh. Technol.}, vol. 73, no. 5, pp. 6212--6224, May 2024.

\bibitem{cite:imagenetc}
D. Hendrycks and T. Dietterich, ``Benchmarking neural network robustness to common corruptions and perturbations,'' in \emph{Proc. Int. Conf. Learn. Represent. (ICLR)}, 2019.

\bibitem{cite:robo3d}
L. Kong, Y. Liu, X. Li, R. Chen, W. Zhang, J. Ren, L. Pan, K. Chen, and Z. Liu, ``Robo3D: Towards robust and reliable 3D perception against corruptions,'' in \emph{Proc. IEEE/CVF Int. Conf. Comput. Vis. (ICCV)}, 2023, pp. 19937--19949.

\bibitem{cite:robobev}
Y. Dong, C. Kang, J. Zhang, Z. Zhu, Y. Wang, X. Yang, H. Su, X. Wei, and J. Zhu, ``Benchmarking robustness of 3D object detection to common corruptions in autonomous driving,'' in \emph{Proc. IEEE/CVF Conf. Comput. Vis. Pattern Recognit. (CVPR)}, 2023, pp. 1022--1032.

\bibitem{cite:transfuser}
K. Chitta, A. Prakash, B. Jaeger, Z. Yu, K. Renz, and A. Geiger, ``TransFuser: Imitation with transformer-based sensor fusion for autonomous driving,'' \emph{IEEE Trans. Pattern Anal. Mach. Intell.}, vol. 45, no. 11, pp. 12878--12895, 2023.

\bibitem{cite:interfuser}
H. Shao, L. Wang, R. Chen, H. Li, and Y. Liu, ``Safety-enhanced autonomous driving using interpretable sensor fusion transformer,'' in \emph{Proc. Conf. Robot Learn. (CoRL)}, 2023, pp. 726--737.

\bibitem{cite:tcp}
P. Wu, X. Jia, L. Chen, J. Yan, H. Li, and Y. Qiao, ``Trajectory-guided control prediction for end-to-end autonomous driving: A simple yet strong baseline,'' in \emph{Adv. Neural Inf. Process. Syst.}, vol. 35, 2022, pp. 6119--6132.

\bibitem{cite:thinktwice}
X. Jia, P. Wu, L. Chen, J. Xie, C. He, J. Yan, and H. Li, ``Think twice before driving: Towards scalable decoders for end-to-end autonomous driving,'' in \emph{Proc. IEEE/CVF Conf. Comput. Vis. Pattern Recognit. (CVPR)}, 2023, pp. 21983--21994.

\bibitem{cite:bevfusion_mit}
Z. Liu, H. Tang, A. Amini, X. Yang, H. Mao, D. L. Rus, and S. Han, ``BEVFusion: Multi-task multi-sensor fusion with unified bird's-eye view representation,'' in \emph{Proc. IEEE Int. Conf. Robot. Autom. (ICRA)}, 2023, pp. 2774--2781.

\bibitem{cite:uniad}
Y. Hu, J. Yang, L. Chen, K. Li, C. Sima, X. Zhu, S. Chai, S. Du, T. Lin, W. Wang, L. Lu, X. Jia, Q. Liu, J. Dai, Y. Qiao, and H. Li, ``Planning-oriented autonomous driving,'' in \emph{Proc. IEEE/CVF Conf. Comput. Vis. Pattern Recognit. (CVPR)}, 2023, pp. 17853--17862.

\bibitem{cite:vad}
B. Jiang, S. Chen, Q. Xu, B. Liao, J. Chen, H. Zhou, Q. Zhang, W. Liu, C. Huang, and X. Wang, ``VAD: Vectorized scene representation for efficient autonomous driving,'' in \emph{Proc. IEEE/CVF Int. Conf. Comput. Vis. (ICCV)}, 2023, pp. 8306--8316.

\bibitem{cite:unibev}
S. Wang, H. Caesar, L. Nan, and J. F. P. Kooij, ``UniBEV: Multi-modal 3D object detection with uniform BEV encoders for robustness against missing sensor modalities,'' in \emph{Proc. IEEE Intell. Veh. Symp. (IV)}, 2024, pp. 2776--2783.

\bibitem{cite:metabev}
C. Ge, J. Chen, E. Xie, Z. Wang, L. Hong, H. Lu, Z. Li, and P. Luo, ``MetaBEV: Solving sensor failures for 3D detection and map segmentation,'' in \emph{Proc. IEEE/CVF Int. Conf. Comput. Vis. (ICCV)}, 2023, pp. 8687--8697.

\bibitem{cite:cafuser}
T. Br\"odermann, C. Sakaridis, Y. Fu, and L. Van Gool, ``CAFuser: Condition-aware multimodal fusion for robust semantic perception of driving scenes,'' \emph{IEEE Robot. Autom. Lett.}, vol. 10, no. 4, pp. 3134--3141, 2025.

\bibitem{cite:crossfuser}
W. Wu, X. Deng, P. Jiang, S. Wan, and Y. Guo, ``CrossFuser: Multi-modal feature fusion for end-to-end autonomous driving under unseen weather conditions,'' \emph{IEEE Trans. Intell. Transp. Syst.}, vol. 24, no. 12, pp. 14378--14392, 2023.

\bibitem{cite:maskfuser}
Y. Duan, X. Guo, Z. Zhu, Z. Wang, Y.-K. Wang, and C.-T. Lin, ``MaskFuser: Masked fusion of joint multi-modal tokenization for end-to-end autonomous driving,'' \emph{arXiv preprint arXiv:2405.07573}, 2024.

\bibitem{cite:policyfuser}
Z. Huang, S. Sun, J. Zhao, and L. Mao, ``Multi-modal policy fusion for end-to-end autonomous driving,'' \emph{Inf. Fusion}, vol. 98, Art. no. 101834, 2023.

\bibitem{cite:kendall_gal}
A. Kendall and Y. Gal, ``What uncertainties do we need in Bayesian deep learning for computer vision?'' in \emph{Adv. Neural Inf. Process. Syst.}, vol. 30, 2017.

\bibitem{cite:gawlikowski_survey}
J. Gawlikowski, C. R. N. Tassi, M. Ali, J. Lee, M. Humt, J. Feng, A. Kruspe, R. Triebel, P. Jung, R. Roscher, M. Shahzad, W. Yang, R. Bamler, and X. X. Zhu, ``A survey of uncertainty in deep neural networks,'' \emph{Artif. Intell. Rev.}, vol. 56, no. 1, pp. 1513--1589, 2023.

\bibitem{cite:michelmore}
R. Michelmore, M. Wicker, L. Laurenti, L. Cardelli, Y. Gal, and M. Kwiatkowska, ``Uncertainty quantification with statistical guarantees in end-to-end autonomous driving control,'' in \emph{Proc. IEEE Int. Conf. Robot. Autom. (ICRA)}, 2020, pp. 7344--7350.

\bibitem{cite:loquercio}
A. Loquercio, M. Segu, and D. Scaramuzza, ``A general framework for uncertainty estimation in deep learning,'' \emph{IEEE Robot. Autom. Lett.}, vol. 5, no. 2, pp. 3153--3160, 2020.

\bibitem{cite:kendall_multitask}
R. Cipolla, Y. Gal, and A. Kendall, ``Multi-task learning using uncertainty to weigh losses for scene geometry and semantics,'' in \emph{Proc. IEEE/CVF Conf. Comput. Vis. Pattern Recognit. (CVPR)}, Salt Lake City, UT, USA, 2018, pp. 7482--7491.

\bibitem{cite:cil}
F. Codevilla, M. M\"uller, A. L\'opez, V. Koltun, and A. Dosovitskiy, ``End-to-end driving via conditional imitation learning,'' in \emph{Proc. IEEE Int. Conf. Robot. Autom. (ICRA)}, 2018, pp. 4693--4700.

\bibitem{cite:cilrs}
F. Codevilla, E. Santana, A. Lopez, and A. Gaidon, ``Exploring the limitations of behavior cloning for autonomous driving,'' in \emph{Proc. IEEE/CVF Int. Conf. Comput. Vis. (ICCV)}, 2019, pp. 9328--9337.

\bibitem{cite:lbc}
D. Chen, B. Zhou, V. Koltun, and P. Kr\"ahenb\"uhl, ``Learning by cheating,'' in \emph{Proc. Conf. Robot Learn. (CoRL)}, 2020, pp. 66--75.

\bibitem{cite:bevfusion_pku}
T. Liang, H. Xie, K. Yu, Z. Xia, Z. Lin, Y. Wang, T. Tang, B. Wang, and Z. Tang, ``BEVFusion: A simple and robust LiDAR-camera fusion framework,'' in \emph{Adv. Neural Inf. Process. Syst.}, vol. 35, 2022, pp. 10421--10434.

\bibitem{cite:pointpainting}
S. Vora, A. H. Lang, B. Helou, and O. Beijbom, ``PointPainting: Sequential fusion for 3D object detection,'' in \emph{Proc. IEEE/CVF Conf. Comput. Vis. Pattern Recognit. (CVPR)}, 2020, pp. 4603--4611.

\bibitem{cite:contfuse}
M. Liang, B. Yang, S. Wang, and R. Urtasun, ``Deep continuous fusion for multi-sensor 3D object detection,'' in \emph{Proc. Eur. Conf. Comput. Vis. (ECCV)}, 2018, pp. 663--678.

\bibitem{cite:mmf}
M. Liang, B. Yang, Y. Chen, R. Hu, and R. Urtasun, ``Multi-task multi-sensor fusion for 3D object detection,'' in \emph{Proc. IEEE/CVF Conf. Comput. Vis. Pattern Recognit. (CVPR)}, 2019, pp. 7337--7345.

\bibitem{cite:drivetransformer}
X. Jia, J. You, Z. Zhang, and J. Yan, ``DriveTransformer: Unified transformer for scalable end-to-end autonomous driving,'' in \emph{Proc. 13th Int. Conf. Learn. Represent. (ICLR)}, 2025.

\bibitem{cite:law}
Y. Li, L. Fan, J. He, Y. Wang, Y. Chen, Z. Zhang, and T. Tan, ``Enhancing end-to-end autonomous driving with latent world model,'' in \emph{Proc. 13th Int. Conf. Learn. Represent. (ICLR)}, 2025.

\bibitem{cite:mcdropout}
Y. Gal and Z. Ghahramani, ``Dropout as a Bayesian approximation: Representing model uncertainty in deep learning,'' in \emph{Proc. Int. Conf. Mach. Learn. (ICML)}, 2016, pp. 1050--1059.

\bibitem{cite:deepensembles}
B. Lakshminarayanan, A. Pritzel, and C. Blundell, ``Simple and scalable predictive uncertainty estimation using deep ensembles,'' in \emph{Adv. Neural Inf. Process. Syst.}, vol. 30, 2017.

\bibitem{cite:edr}
A. Amini, W. Schwarting, A. Soleimany, and D. Rus, ``Deep evidential regression,'' in \emph{Adv. Neural Inf. Process. Syst.}, vol. 33, 2020, pp. 14927--14937.

\bibitem{cite:edl}
M. Sensoy, L. Kaplan, and M. Kandemir, ``Evidential deep learning to quantify classification uncertainty,'' in \emph{Adv. Neural Inf. Process. Syst.}, vol. 31, 2018.

\bibitem{cite:ddu}
J. Mukhoti, A. Kirsch, J. van Amersfoort, P. H. S. Torr, and Y. Gal, ``Deep deterministic uncertainty: A new simple baseline,'' in \emph{Proc. IEEE/CVF Conf. Comput. Vis. Pattern Recognit. (CVPR)}, 2023, pp. 24384--24394.

\bibitem{cite:filos}
A. Filos, P. Tigkas, R. McAllister, N. Rhinehart, S. Levine, and Y. Gal, ``Can autonomous vehicles identify, recover from, and adapt to distribution shifts?'' in \emph{Proc. Int. Conf. Mach. Learn. (ICML)}, 2020, pp. 3145--3153.

\bibitem{cite:lidar_fog}
M. H\"ahner, C. Sakaridis, D. Dai, and L. Van Gool, ``Fog simulation on real LiDAR point clouds for 3D object detection in adverse weather,'' in \emph{Proc. IEEE/CVF Int. Conf. Comput. Vis. (ICCV)}, 2021, pp. 15263--15272.

\bibitem{cite:lidar_snow}
M. H\"ahner, C. Sakaridis, M. Bijelic, F. Heide, F. Yu, D. Dai, and L. Van Gool, ``LiDAR snowfall simulation for robust 3D object detection,'' in \emph{Proc. IEEE/CVF Conf. Comput. Vis. Pattern Recognit. (CVPR)}, 2022, pp. 16343--16353.

\bibitem{cite:carla_gear}
F. Nesti, G. Rossolini, G. D'Amico, A. Biondi, and G. Buttazzo, ``CARLA-GeAR: A dataset generator for a systematic evaluation of adversarial robustness of deep learning vision models,'' \emph{IEEE Trans. Intell. Transp. Syst.}, vol. 25, no. 8, pp. 9840--9851, 2024.

\bibitem{cite:resilient}
K. Park, Y. Kim, D. Kim, and J. W. Choi, ``Resilient sensor fusion under adverse sensor failures via multi-modal expert fusion,'' in \emph{Proc. IEEE/CVF Conf. Comput. Vis. Pattern Recognit. (CVPR)}, 2025, pp. 6720--6729.

\bibitem{cite:regnety}
I. Radosavovic, R. P. Kosaraju, R. Girshick, K. He, and P. Doll\'ar, ``Designing network design spaces,'' in \emph{Proc. IEEE/CVF Conf. Comput. Vis. Pattern Recognit. (CVPR)}, 2020, pp. 10425--10433.

\bibitem{cite:centernet}
X. Zhou, D. Wang, and P. Kr\"ahenb\"uhl, ``Objects as points,'' \emph{arXiv preprint arXiv:1904.07850}, 2019.

\bibitem{cite:focal}
H. Law and J. Deng, ``CornerNet: Detecting objects as paired keypoints,'' \emph{Int. J. Comput. Vis.}, vol. 128, no. 3, pp. 642--656, 2020.

\bibitem{cite:gru}
K. Cho, B. van Merri{\"e}nboer, C. Gulcehre, D. Bahdanau, F. Bougares, H. Schwenk, and Y. Bengio, ``Learning phrase representations using RNN encoder--decoder for statistical machine translation,'' in \emph{Proc. Conf. Empirical Methods Natural Lang. Process. (EMNLP)}, Doha, Qatar, 2014, pp. 1724--1734.

\end{thebibliography}
\end{document}